# Prompts Matter: Comparing ML/GAI Approaches for Generating Inductive Qualitative Coding Results

*John Chen, Alexandros Lotsos, Lexie Zhao, Grace Wang, Uri Wilensky, Bruce Sherin, Michael Horn*

*Northwestern University*

## Abstract

Inductive qualitative methods have been a mainstay of education research for decades, yet it takes much time and effort to conduct rigorously. Recent advances in artificial intelligence, particularly with generative AI (GAI), have led to initial success in generating inductive coding results. Like human coders, GAI tools rely on instructions to work, and how to instruct it may matter. To understand how ML/GAI approaches could contribute to qualitative coding processes, this study applied two known and two theory-informed novel approaches to an online community dataset and evaluated the resulting coding results. Our findings show significant discrepancies between ML/GAI approaches and demonstrate the advantage of our approaches, which introduce human coding processes into GAI prompts.

## Introduction

Inductive qualitative coding, the process of identifying and generating potential themes or codes from data, has been widely adopted by many fields, including education, to create new theoretical insights that help us understand the nuances of human interactions. However, as qualitative researchers know, inductive coding can be highly time-consuming (Kalman, 2019), and it is challenging to meet theoretical expectations. While prior studies have leveraged machine learning (ML) approaches to support qualitative analysis, fewer have touched on inductive coding. Recently, researchers have successfully used Generative AI (GAI) to generate inductive coding results. GAI requires instructions just like humans, with researchers noting the strong impact of instructions on GAI outcomes. However, early GAI studies on inductive coding mostly worked with small individual pieces of data and/or codebooks, making it difficult to compare them effectively.

We present an exploratory study that systematically and automatically applies two known ML/GAI approaches to generate inductive codes for an online community dataset. Building on them, we proposed two novel, theory-informed approaches that identified more codes and nuanced insights from the same dataset. We evaluated four generation results regarding size, groundedness, overly broad, and overlapping concepts. Our findings show significant discrepancies between different ML/GAI approaches, suggesting the advantage of introducing human coding processes into GAI prompts. The large number and prevalence of overlapping codes suggest the urgency of our follow-up work, a human-AI approach for measuring inductive qualitative coding results.

---

*The paper itself is 6 pages long with 1992 words. The rest is appendixes containing the original data.*

## Background

Qualitative analysis theorists from both thematic analysis and grounded theory suggest inductive coding as a key approach to discovering and constructing concepts from a dataset (J. M. Corbin & Strauss, 1990; Fereday & Muir-Cochrane, 2006). Yet, as an open-ended process, inductive coding is suspect to rigor issues (Roberts et al., 2019) and can be very time-consuming (Deterding & Waters, 2021). To do it rigorously, grounded theory asks for line-by-line or incident-by-incident coding, looking for all possible interpretations while grounded in the dataset (Willig & Rogers, 2017). However, the expectations could be challenging to match. Thematic analysis is more flexible but also asks for groundedness, i.e., whether the codes and themes are derived from and closely match the data (Braun & Clarke, 2012).

Researchers have incorporated various ML/AI methods to help alleviate some of the issues with inductive coding. For example, topic modeling (Wallach, 2006), an unsupervised ML technique to identify groups of semantically similar words (i.e., topics), has been used to analyze qualitative survey data (Baumer et al., 2017), social media posts, and online discussion board data (Saravani et al., 2023), etc. The topics identified were used to focus the researcher's attention on important parts of the dataset for deeper analysis and to automatically code the dataset in preparation for future rounds of analysis. However, as an unsupervised technique, the output of topic modeling can be difficult for researchers to interpret, measure, or evaluate despite previous work, making the method inaccessible to many researchers (e.g., Grootendorst, 2022; Sievert & Shirley, 2014).

Recently, researchers have found some success in using GAI for inductive coding. To use LLMs for inductive coding, researchers supply a piece of data with relevant instructions (e.g., research questions, coding instructions, desired output format), and the LLM would respond accordingly. Repeating this process, researchers could consistently produce codes that humans found more interpretable and useful (e.g., DePaoli, 2023; Sinha et al., 2024). However, LLMs could still lack nuance and fail to grasp less linguistically straightforward themes (DePaoli, 2023; Hamilton et al., 2023). LLMs could produce results not grounded in the data (Byun et al., 2023; De Paoli, 2023), or remain on a coarser grain size of analysis than desired (Sinha et al., 2024; Zambrano et al., 2023). Studies highlight that careful prompt design and thoughtful prompting strategies are crucial to addressing these limitations when using LLMs for inductive coding (Byun et al., 2024; Sinha et al., 2024). As such, we pose the following research questions:

- What codes can we discover from text-based datasets using existing ML/GAI methods?
- How can we do so more effectively?

## Context and Method

We developed a system that automatically and systematically runs two known and two novel approaches on text-based datasets for inductive coding. Though we tested on many LLMs, due to page constraints, we only report the results of GPT-4o, one of the most powerful models in 2024.

To understand how LLMs could identify emergent insights from unstructured datasets, we used a conversational dataset from Physics Lab's online message groups. The original research question was to understand how Physics Lab's online community emerges, which is an open-ended question suitable for an inductive qualitative analysis. The example dataset covers 127 messages between designers and teachers over the first two months of the community. We separated the conversation flow into chunks for human and machine coders with signal processing techniques (*Appendix 1*). Four coders independently open-coded the dataset before LLMs. Two of them have worked with the community for years, bringing in-depth knowledge about the conversations; the other two have never worked with the community, bringing fresh perspectives with little preconceptions.

We assigned the same roles and tasks to machine coders, such as "You are an expert in thematic analysis with grounded theory, working on open coding." We also provide the research question and dataset information to human and machine coders (See *Appendix 2*). Then, we instruct each machine coder to follow one approach, such as identifying labels from a list of quotes (as in Topic Modeling). To ensure the process was truly inductive, machine coders received no codebooks or examples beforehand. Due to differences in datasets/research contexts, replicating ML/GAI methods from prior studies is challenging. In *Appendix 3*, we provide the original LLM prompts and our replication side-by-side to help with transparency.

Two authors counted the AI-generated codes and flagged noticeable findings while reading each codebook independently. They independently marked potential groundedness issues (i.e., when humans can not reasonably connect the label and the data) and overly broad codes (i.e., when the code provides little information to the researchers) and reconciled their differences. To avoid selection bias, we provide the four codebooks and examples in *Appendix 4.*

## Study Results

|  | Topic Modeling + LLM | Chunk-Level LLM Coding | Item-Level LLM Coding | Item-Level Coding w/ Verb Phrases |
|---|---|---|---|---|
| # of codes | 23 | 48 | 240 | 271 |
| # of groundedness issues | 2 | 1 | 2 | 0 |
| # of overly broad | 2 | 3 | 5 | 7 |

Table 1. An overview of our human evaluation results.

| LLM Higher Level, Themes | LLM Lower Level, Tags | LLM Lower Level, Verb Phrases | 4 Human Coders |
|---|---|---|---|
| community feedback | comparative feedback | acknowledge feedback | appreciation of feedback |

| community feedback loop | positive feedback | address feedback process | community feedback |
|---|---|---|---|
| encouragement of user feedback | user feedback | consider user feedback | eliciting feedback |
| user feedback | user feedback request | encourage community feedback | encouraging feedback |
| user feedback and suggestions | user feedback response | invite feedback | invite for feedback |
|  | user feedback solicitation | plan to gather feedback | justified feedback |
|  |  | provide feedback | positive feedback |
|  |  | provide positive feedback | prompting user feedback |
|  |  | provide specific feedback | reaction to feedback |
| **BERTopic, Topics** |  | solicit user feedback | response to feedback |
| iterative development based on user feedback |  |  | soliciting feedback |
| user feedback and communication |  |  | taking feedback |
|  |  |  | user experience feedback |

Table 2. Comparing different approaches' results around the concept of "feedback."

## Topic Modeling

While earlier studies manually interpret the topics (Baumer et al., 2017), a more recent method, BERTopic, uses LLMs to explain the topic modeling results for better human interpretability. Since BERTopic's prompt intends for general-purpose label-making, we adapted and optimized it (*Appendix 3.1*), similar to a recently reported method (Katz et al., 2024). Yet, we found the approach only capable of finding content topics and surface-level interpretations.

This method resulted in 23 codes (*Appendix 4.1*), most related to content topics from the dataset, e.g., "software update process" or "user guidance and instructions." We marked two codes as problematic in groundedness. For both, the clustering algorithm identified too many examples (e.g., one has 34 examples, 26% of the dataset) for LLMs to find an appropriate label. Occasionally, BERTopic made superficial-level interpretations, e.g., the "informal interaction" code included cases of "Hello," "Sorry," or "Yes, yes." Some labels overlap conceptually, e.g., the "software update process" and "software updates and downloads."

## Chunk-level LLM Coding

For the next approach, we replicated a well-documented example from AIED 2024 (Barany et al., 2024), which asked LLMs to provide codebooks for each chunk of an interview. Many more recent studies followed the approach as well (Parfenova, 2024). We found it capable of identifying more insights from the data, had few groundedness issues, yet suffered from overly broad and overlapping codes.

The original prompt is unstructured, resulting in different response formats that require humans to work on. To avoid introducing bias, after discussing with one author of the paper, we automated the process by structuring the prompt (*Appendix 3.2*) and merging results only when labels were precisely the same.

This method resulted in 48 codes (*Appendix 4.2*), twice as many as BERTopic. Rather than simply describing what happened, it found codes related to the conversation process, such as "role identification" or "recognition of effort," which were missing from the previous approach. As the number increased, the codebook also had more overlapping codes: e.g., "community feedback" is conceptually close to "community feedback loop"; "resource sharing" is the same as "sharing resources;" or codes that included one another, e.g., "user feedback" includes "user feedback and impact." Finally, it sometimes brought up very broad topics, such as "community engagement" and "user engagement" that could include everything in the dataset.

## Item-level LLM Coding

Inspired by grounded theorists who suggest line-by-line coding (Gibbs, 2007) and a recent study that asked GPT-4 to find a single code for each item (Sinha et al., 2024), we propose a novel approach that instructs LLMs to write a plan before coding, then identify multiple codes for each item (prompt in *Appendix 3.3*). This approach seems more capable of identifying nuances, although overly broad and overlapping codes persist.

Our item-level approach identifies 240 codes, almost six times as many as the chunk-level approach (*Appendix 4.3*), while also more easily identifying nuances across the text. For example, it identifies nine codes around the concept of "software update," a common topic in the dataset. It presented a much more detailed perspective than previous approaches, which only identified "update announcement" and "update planning." However, it still generates overly broad concepts such as "user feedback," or overlapping ones such as "update inquiry" and "update status inquiry."

We found this approach impressively capable of reading contexts. When it coded "User: Yes" as "professional engagement," it seemed to learn from a previous message, where a designer asked for the same user's identity. However, such capability may become a negative asset. In the only problematic case for groundedness (0.83%), GPT-4o labelled a message as "new user interaction." It was indeed the user's first interaction, yet the input data did not include that information.

## Item-level Coding w/ Verb Phrases

Inspired by a study using grounded theory analysis (Davis et al., 2020), we improved the item-level approach by asking LLMs to use verb phrases as labels, hoping to distinguish the nuances at the level of micro-actions (*Appendix 4.4*). The revised approach found 271 codes, a 13% increase, while showing substantial qualitative improvement by eliminating many overly broad codes and identifying more nuances.

Instead of vague codes such as "user feedback," the Verb Phrases approach identified ten sub-level codes such as "acknowledge feedback" or "invite feedback," more than any other approach we have tested (*Appendix 6* for a comparison) or any individual human coders. Still, the number is less than four humans combined. We identified no groundedness issues. Still, we found four overly broad codes (2.58%), all related to "community engagement" in different forms - a common theme in the dataset.

## Discussions

Our study reveals clear qualitative differences between codebooks generated from four different ML/GAI approaches. The item-level LLM coding approach with the usage of verb phrases performed the best (Table 1), with a considerable capability of identifying nuanced insights while reducing issues with groundedness or vagueness. Our finding showcases 1) the importance of prompts in using ML/GAI to generate inductive coding results and 2) the potential of applying human qualitative coding processes in GAI prompts.

Except for the BERTopic approach, we found few issues of groundedness. LLMs did not always use optimal labels, but they rarely hallucinated labels without connection to the data. One reason is that the LLM was explicitly instructed to ground its generation in the input data, which has been found to reduce hallucinations (e.g., Lewis et al., 2020). Another reason is the nature of inductive coding. An interpretation or construction from the underlying data may be "shallow," "irrelevant," or "incomplete," but can hardly be "wrong" in a strict sense (Terry et al., 2017).

It is essential to discuss the implications of our work. For qualitative researchers, inductive coding means more than producing the codes; the process enables them to get closer to the underlying dataset (J. Corbin & Strauss, 2008). From this epistemological standpoint, it seems that machines can only imitate the results of open coding processes. Even when our Item-Level Verb Phrases approach identified codes close to human results, claiming that it (with GPT-4o) can closely imitate human results would be far-reaching. To rigorously claim that requires working with more research contexts. Moreover, none of the approaches fully eliminated overly broad or overlapping codes, which may be attributed to the stochastic nature of LLMs. With hundreds of codes and the ambiguity of human language (e.g., could we claim that "soliciting feedback" covers the same idea as "eliciting feedback" ), examining where LLMs or humans *missed* was impractical to do by hand. On the other hand, it also provides rich opportunities for qualitative researchers to identify novel insights from machine coders - particularly when machines identified extra codes than humans.

A rigorous, computational(-assisted) approach to evaluating qualitative codes is urgently needed. While consistency is widely used to assess deductive coding, it can only address issues *after* establishing the codebook. Early attempts using consistency or coverage to evaluate codebooks (Zhao et al., 2024) suffer a similar issue: using human results as the "golden standard" for machines to match, they miss opportunities for machine approaches to provide novel insights. Hence, a theory-informed human-AI approach becomes necessary as the boundary between "right" and "wrong" is blurry for inductive coding. With that in mind, we will soon present our follow-up work on systematic measurement of inductive coding results.

*Appendix 1: An Example Chunk of Our Working Dataset*

| ID | Anonymized Usernames | Time | Message Content |
|---|---|---|---|
| 2-55 | Designer | 11/06 10:04 | We will also update the multimeter and powered solenoid this week. If there are any other features, characteristics, or electronic components you hope to include in the production plan, please leave a message in the group, and we will consider it as much as possible. The previously mentioned feature of saving experiments to the cloud/local is also in the plan. Thank you, everyone! |
| 2-56 | Ficta McFiction | 11/06 10:10 | [Emoji] |
| 2-57 | Sam Sampleton | 11/06 10:11 | [Emoji][Emoji] |
| 2-58 | Testy McTestface | 11/14 10:23 | Hello everyone |
| 2-59 | Designer | 11/14 10:26 | Hello :) |
| 2-60 | Ficta McFiction | 11/14 10:30 | Are there any other updates recently? |
| 2-61 | Designer | 11/14 10:30 | [Image] Updates in preparation |
| 2-62 | Designer | 11/14 10:30 | This is quite complex, so it will take more time... Hopefully, it can be released this week |
| 2-63 | Designer | 11/14 10:31 | There will be: multimeter; powered solenoid; semiconductor capacitor; support for conversion to ideal ammeter (more convenient for problem-solving and middle school teaching) |
| 2-64 | Testy McTestface | 11/14 10:32 | Is there a user manual? |
| 2-65 | Designer | 11/14 10:32 | What problems did you encounter during use? |
| 2-66 | Designer | 11/14 10:32 | We try to design it so that it can be used without additional instructions, or we will provide some prompts when you open it for the first time based on feedback |
| 2-67 | Testy McTestface | 11/14 10:33 | Mainly, the school is building an information-based school |
| 2-68 | Testy McTestface | 11/14 10:33 | Need the user manual for this software |
| 2-69 | Designer | 11/14 10:34 | Oh, please message me privately and tell me what you roughly need |
| 2-70 | Sam Sampleton | 11/14 10:34 | This is good, I hope it can be very convenient to draw circuit diagrams, and I also hope there is a function to hide the background grid with one click, which is convenient for us teachers to take screenshots for test papers. Currently, there is no software that makes it easy to draw circuit diagrams for exam questions |
| 2-71 | Designer | 11/14 10:34 | Actually, we are researching the bidirectional conversion between physical diagrams and circuit diagrams |
| 2-72 | Designer | 11/14 10:34 | Of course, manual editing will also be allowed, but it might be a bit later |
| 2-73 | Testy McTestface | 11/14 10:35 | Can you also include mechanics experiments? |
| 2-74 | Designer | 11/14 10:35 | For example, you can see the corresponding circuit diagram after connecting the physical diagram, or vice versa |
| 2-75 | Designer | 11/14 10:35 | Mechanics will have to wait until electromagnetism is figured out; it will take some more time |
| 2-76 | Sam Sampleton | 11/14 10:35 | Oh, this is nice |
| 2-77 | Designer | 11/14 10:35 | Hope to figure out electromagnetism before the end of the year |
| 2-78 | Testy McTestface | 11/14 10:35 | The 3D effect of your software is very good |
| 2-79 | Designer | 11/14 10:35 | With bidirectional conversion, you can directly do problems in the application |
| 2-80 | Ficta McFiction | 11/14 10:36 | It's already starting to take shape [Emoji] |
| 2-81 | Designer | 11/14 10:36 | For example, see the circuit diagram to connect the physical diagram, or vice versa |
| 2-82 | Sam Sampleton | 11/14 10:37 | Don't aim for completeness, it should be categorized and refined one by one |
| 2-83 | Designer | 11/14 10:37 | Hmm. |
| 2-84 | Designer | 11/14 10:38 | Physics mainly focuses on electricity and mechanics; other directions don't have much room for free experiments before high school. In the future, we might do some demonstration experiments based on everyone's needs, but it won't be as open as electricity |
| 2-85 | Sam Sampleton | 11/14 10:39 | Indeed |
| 2-86 | Designer | 11/14 10:39 | I myself have some interest in chemistry |
| 2-87 | Designer | 11/14 10:39 | We'll see if there's a suitable opportunity to do one later |

| 2-88 | Designer | 11/14 10:40 | For electricity, we plan to create a function for simulated experiment assessment; it will add many experimental details, such as zero adjustment (of course, only useful in specific modes, otherwise it would be cumbersome to use regularly) |
|---|---|---|---|
| 2-89 | Sam Sampleton | 11/14 10:41 | This idea is really good |
| 2-90 | Designer | 11/14 10:49 | When we are ready to start, we will consult everyone in the group |

*Appendix 2: Contextual Information on the Research Question and Dataset*

| Research Question | How did Physics Lab's online community emerge? | |
|---|---|---|
| **Dataset** | Size | 127 lines |
| | Dated | 2017-10 ~ 2017-12 |
| | Language | Chinese translated to English with gpt-4o, manually verified. |
| | Processing Method | Automatically chunked using signal processing techniques. To mitigate the impacts of automatic chunking, for each chunk (conversation), we fetched 3 messages from the previous one and 3 from the next one. |

## Appendix 3: Prompting Templates for Each Approach

### 3.1 Topic Modeling + LLM Prompts

| BERTopic Original Prompt | Revised Prompt |
|---|---|
| " I have a topic that contains the following documents: *{Documents}*.<br>The topic is described by the following keywords: '*{Keywords}*'.<br>Based on the above information, can you give a short label on the topic? " | System Prompt:<br>" You are an expert in thematic analysis with grounded theory, working on open coding.<br>You identified a topic from the input quotes. Each quote is independent from another.<br>*${ResearchQuestion}*<br>*${CodingNotes}*<br><br>Always follow the output format:<br>===<br>Thought: {What is the most common theme among the input quotes? Do not over-interpret the data.}<br>Label: {A single label that faithfully describes the topic}<br>=== " |
| | User Prompt:<br>" Quotes: *{Documents}*<br>Keywords: *{Keywords}* " |

### 3.2 Chunk-Level LLM Coding Prompts

| Unstructured AIED Paper Prompt | Revised Structured Prompt |
|---|---|
| " Hi ChatGPT, I want to analyze the following interaction between an instructor and some students:<br>*{Conversation}*<br>Please give me a codebook to analyze the instructional methodologies and the sentiment within this interaction. " | System Prompt:<br>"Hi ChatGPT, I want to analyze the following interaction in one of Physics Lab's online message groups.<br>Please give me a codebook to analyze factors within this interaction that could contribute to the research.<br>*${ResearchQuestion}*<br>*${CodingNotes}*<br>For each code, try to find 3 quotes. Always follow the output format:<br>===<br>## Label: A label of code 1<br>Definition: A definition of code 1<br>- "Example quote 1"<br>- "Example quote 2"<br>## ... " |
| | User Prompt:<br>"*{Conversation}*" |

## 3.3 Item-Level LLM Coding Prompts

| System Prompt |
|---|

> " You are an expert in thematic analysis with grounded theory, working on open coding.
> Your goal is to identify multiple low-level tags for each message.
> When writing tags, balance between specifics and generalizability across messages.
> *${ResearchQuestion}*
> *${CodingNotes}*
>
> Always follow the output format:
> ===
> Thoughts: {A paragraph of plans and guiding questions about analyzing the conversation from multiple theoretical angles}
> Tags for each message (*${Messages.length}* in total):
> 1. tag 1; tag 2; tag 3...
> ...
> *${Messages.length}*. tag 4; tag 5; tag 6...
> Summary: {A somehow detailed summary of the conversation, including previous ones}
> Notes: {Notes and hypotheses about the conversation until now}         "

## 3.4 Item-Level Coding w/ Verb Phrases

We slightly edited the prompt so LLMs would conduct open-coding with verb phrases. We only made two changes: first, "Always use verb phrases" in the system prompt; second, in the output template, we changed "tags" to "interpretations" and "tag" to "phrase". Interestingly, LLMs often failed to use verb phrases if "tags" were kept unchanged.

## Appendix 4: Four Generated Codebooks

### 4.1 Topic-Modeling Codes

| Label | Examples |
|-------|----------|
| circuit diagram tool development | ● 2-70: User: This is good, I hope it can be very convenient to draw circuit diagrams, and I also hope there is a function to hide the background grid with one click, which is convenient for us teachers to take screenshots for test papers. Currently, there is no software that makes it easy to draw circuit diagrams for exam questions<br>● 2-71: Designer: Actually, we are researching the bidirectional conversion between physical diagrams and circuit diagrams<br>● 2-74: Designer: For example, you can see the corresponding circuit diagram after connecting the physical diagram, or vice versa<br>● 2-81: Designer: For example, see the circuit diagram to connect the physical diagram, or vice versa<br>● 2-91: Designer: Consulting the teachers in the group: which type of intersection is used in the circuit diagrams in the current textbooks? [Image]<br>● 2-101: Designer: This version of the circuit diagram is for testing only... You can drag the Editor because the auto-layout algorithm is not very stable and may produce some odd results |
| emoji communication | ● 2-56: User: [Emoji]<br>● 2-57: User: [Emoji][Emoji]<br>● 2-80: User: It's already starting to take shape [Emoji]<br>● 2-104: User: Will there be an update this week? [Emoji]<br>● 2-107: User: [Emoji]<br>● 2-108: User: [Emoji]<br>● 2-116: User: [Emoji]<br>● 2-119: User: [Emoji]<br>● 2-121: User: [Emoji] |
| experimental simulations in electricity | ● 2-84: Designer: Physics mainly focuses on electricity and mechanics; other directions don't have much room for free experiments before high school. In the future, we might do some demonstration experiments based on everyone's needs, but it won't be as open as electricity<br>● 2-88: Designer: For electricity, we plan to create a function for simulated experiment assessment; it will add many experimental details, such as zero adjustment (of course, only useful in specific modes, otherwise it would be cumbersome to use regularly) |
| feature prioritization | ● 2-9: Designer: Hmm... We will probably prioritize completing the electrical section first, then magnetism, and then other parts.<br>● 2-10: Designer: Before starting mechanics, we will gather opinions again~ Otherwise, I'm afraid I won't remember everything. |

| Label | Examples |
|---|---|
| feature prioritization | • 2-12: User: [Emoji] Hello everyone, may I ask where I can download the PC / interactive whiteboard version of Physics Lab? |
| | • 2-13: User: I saw the group files, thank you. |
| | • 2-16: Designer: The student power supply supports both DC and AC and is an ideal component. |
| | • 2-17: Designer: Okay. Please give more suggestions! |
| | • 2-22: User: Win7 should be fine. |
| | • 2-23: Designer: 7 is okay. |
| | • 2-30: Designer: New Features |
| | Electronic components will be damaged after a short process, rather than immediately. |
| | Clearing the desktop will now display a confirmation interface. |
| | Supports undoing the creation and deletion of wires and components. |
| | Appliances now display the effective value of alternating current. |
| | (PC) You can now exit the application using the Esc key. |
| | |
| | New Components |
| | Added a sensitive ammeter. |
| | Added a student power supply (ideal AC/DC power supply). |
| | |
| | Adjustments to Component Properties |
| | Batteries now have adjustable internal resistance and are no longer ideal power sources. |
| | Incandescent bulbs now have volt-ampere characteristic parameters and are no longer ideal resistors. |
| | The resistance law experimenter now uses real formulas for calculations, with adjustable parameters. |
| | Hidden terminal blocks 3 and 4. |
| | |
| | Bug Fixes |
| | Fixed an issue with unit conversion in Editor properties. |
| | Terminal arrows no longer show jumping animations. |
| | There may have been calculation errors with certain circuit connections. |
| | • 2-46: User: The simulation effect of this software is really good. When I used it in class yesterday, the students were amazed. |
| | • 2-49: Designer: Version 1.0.1 of Physics Lab. It is expected to take another one or two days in the Apple market (waiting for review). |
| | • 2-53: Designer: Group sharing. |
| | • 2-60: User: Are there any other updates recently? |
| | • 2-62: Designer: This is quite complex, so it will take more time... Hopefully, it can be released this week |
| | • 2-64: User: Is there a user manual? |
| | • 2-66: Designer: We try to design it so that it can be used without additional instructions, or we will provide some prompts when you open it for the first time based on feedback |

| Label | Examples |
|---|---|
| feature prioritization | • 2-69:    Designer: Oh, please message me privately and tell me what you roughly need<br>• 2-72:    Designer: Of course, manual editing will also be allowed, but it might be a bit later<br>• 2-75:    Designer: Mechanics will have to wait until electromagnetism is figured out; it will take some more time<br>• 2-77:    Designer: Hope to figure out electromagnetism before the end of the year<br>• 2-79:    Designer: With bidirectional conversion, you can directly do problems in the application<br>• 2-83:    Designer: Hmm.<br>• 2-86:    Designer: I myself have some interest in chemistry<br>• 2-90:    Designer: When we are ready to start, we will consult everyone in the group<br>• 2-93:    User: Yes, the common one is still the old style<br>• 2-100:  Designer: So, use both?<br>• 2-103:  Designer: Thank you all for your support. We will do better!<br>• 2-106:  Designer: The update is quite large...<br>• 2-109:  Designer: Don't worry, don't worry, it will come, just optimizing the circuit diagram one last time<br>• 2-110:  Designer: Although you can manually adjust the layout<br>• 2-112:  User: Yes, the spirit of craftsmanship that strives for perfection 🤭<br>• 2-115:  Designer: v1.0.2 - 17/11/29<br>        # New Features<br>        Multi-language support: Japanese, German, French.<br>        Preliminary support for converting circuit components to circuit diagrams.<br>        Support for augmented reality mode on the latest iOS devices.<br>        More realistic 3D appearance for various meters.<br><br>        # Update Preview<br>        The next version will provide multimeter and meter zeroing.<br>        The next version will provide AC waveform display functionality.<br>        The next version will provide capacitors and energized solenoids.<br><br>        # Component Property Adjustments<br>        Dry cells can now be ideal power sources.<br>        All meters can now switch between ideal/realistic modes.<br><br>        # Special Tips<br>        The Esc shortcut key will now minimize the window.<br>        The Ctrl+E shortcut key can switch between exam and normal modes.<br>• 2-117:  Designer: The new version has already been sent in the group<br>• 2-118:  Designer: iOS is waiting for review~ |

| Label | Examples |
|---|---|
| feature requests for physics experiments | • 2-8:    User: First, let's pay homage to the experts, then I'll make a small request. Could you create a dynamic demonstration of mechanical waves and mechanical vibrations, such as the propagation of transverse and longitudinal waves, wave superposition, diffraction, and interference? Also, for optical experiments, it would be great to have optical benches, single slits, double slits, and polarizers to demonstrate optical experiments.<br>• 2-38:    User: If there could be an export function, or the ability to save or import experiments, it would be convenient. We could set up the parameters in the office and directly import them in class.<br>• 2-73:    User: Can you also include mechanics experiments? |
| future planning and development | • 2-39:    Designer: There will be.<br>• 2-87:    Designer: We'll see if there's a suitable opportunity to do one later<br>• 2-95:    Designer: We are working on this part |
| image and avatar management | • 2-0:    Designer: [Image]<br>• 2-1:    Designer: @Morning Tea Moonlight How can I upload a high-definition, uncensored version of this crappy avatar?<br>• 2-14:    Designer: [Image]<br>• 2-19:    User: [Image]<br>• 2-61:    Designer: [Image] Updates in preparation<br>• 2-92:    Designer: [Image] This one, right?<br>• 2-102:    Designer: [Image] |
| informal interaction | • 2-37:    User: @Designer Yes, yes.<br>• 2-42:    Designer: 😂<br>• 2-50:    Designer: Sorry for the late-night disturbance :)<br>• 2-59:    Designer: Hello :) |
| insufficient data | • 2-45:    Designer: Hmm hmm.<br>• 2-97:    Designer: Okay. |
| interface layout decisions | • 2-94:    Designer: Uh... left side or right side<br>• 2-96:    User: Left side<br>• 2-99:    User: But it's better to use the right side for non-crossing |
| iterative development based on user feedback | • 2-15:    Designer: From the next update, dry batteries will no longer be ideal components (you can remove the internal resistance to simulate).<br>• 2-55:    Designer: We will also update the multimeter and powered solenoid this week. If there are any other features, characteristics, or electronic components you hope to include in the production plan, please leave a message in the group, and we will consider it as much as possible. The previously mentioned feature of saving experiments to the cloud/local is also in the plan. Thank you, everyone! |

| Label | Examples |
|---|---|
| | • 2-63:   Designer: There will be: multimeter; powered solenoid; semiconductor capacitor; support for conversion to ideal ammeter (more convenient for problem-solving and middle school teaching) |
| on-screen keyboard functionality | • 2-31:   User: It is recommended to add an exit button function to the PC version. Many regions now use all-in-one touch screen machines without physical keyboards.<br>• 2-33:   Designer: Speaking of which, doesn't that mean every place where numbers are input should have a soft keyboard?<br>• 2-34:   Designer: Touch screens have their own on-screen keyboards.<br>• 2-35:   Designer: But you can't bring it up without touching the input, that's probably the case. |
| pc version inquiries | • 2-2:   User: PC version?<br>• 2-18:   User: How do you use the PC version? |
| simplification and refinement in design | • 2-82:   User: Don't aim for completeness, it should be categorized and refined one by one<br>• 2-98:   Designer: Make it simpler...<br>• 2-111:   Designer: It's still better to make the auto-generated one as good as possible<br>• 2-113:   Designer: This belongs to the kind of feature that, once done, will ensure long-term stability... Adding various components is actually simpler |
| software update process | • 2-32:   Designer: Hmm... I'll add it in the next update. PC updates are usually the fastest, so there will probably be another round this weekend.<br>• 2-123:   Designer: The Android version is expected to update tonight |
| software updates and downloads | • 2-6:   Designer: Hello everyone~ The development plan and PC download address are in the group announcement.<br>• 2-29:   Designer: Hello everyone~ Version 1.01 has been released in the group files, the updates are as follows:<br>• 2-43:   Designer: It seems that local export needs to be supported.<br>• 2-54:   Designer: You can download Windows in the group files.<br>• 2-122:   Designer: The one in the group is the PC version |
| system compatibility | • 2-20:   Designer: It does not support the XP system.<br>• 2-27:   Designer: XP compatibility might need to be looked into later... probably need to install a virtual machine.<br>• 2-28:   Designer: Theoretically, it should be compatible (but the machine itself shouldn't be too old, probably from 2008 onwards). |

| Label | Examples |
|---|---|
| technical and infrastructural challenges in educational settings | ● 2-3:   Designer: I'll upload one now... Are you a teacher?<br>● 2-21:   User: Dizzy, the computer system for teachers at our school is quite old.<br>● 2-36:   Designer: Last time I tried it in middle school, there's a trick where you click the link button in the top right, and when the browser pops up, it's no longer full screen... = =<br>● 2-40:   Designer: Does the class have internet?<br>● 2-41:   User: Generally not. Ever since an adult image popped up during a major city-level open class, the school has disabled the network on classroom computers [Emoji].<br>● 2-44:   User: Most schools will disable the network on classroom computers to prevent students from going online or to avoid various software auto-downloads from slowing down the computers.<br>● 2-47:   Designer: Haha. Are you a middle school or high school teacher?<br>● 2-48:   User: I taught high school for seven years, and now I've been teaching middle school for eight years.<br>● 2-67:   User: Mainly, the school is building an information-based school |
| user feedback and communication | ● 2-7:   Designer: If you have any suggestions or requirements, feel free to bring them up.<br>● 2-25:   Designer: No need to be polite, if you encounter any problems during use, you can directly mention them in the group.<br>● 2-65:   Designer: What problems did you encounter during use? |
| user guidance and instructions | ● 2-51:   User: How to download?<br>● 2-68:   User: Need the user manual for this software |
| user interaction and gratitude | ● 2-4:   User: Yes.<br>● 2-11:   User: Okay, okay~ Thank you for your hard work.<br>● 2-24:   User: Thank you.<br>● 2-52:   User: Excuse me.<br>● 2-58:   User: Hello everyone<br>● 2-76:   User: Oh, this is nice<br>● 2-85:   User: Indeed<br>● 2-89:   User: This idea is really good<br>● 2-120:   User: ok |
| user satisfaction with software features | ● 2-26:   User: This software is great! It's quite practical, unlike some software that tries to be comprehensive but ends up being inconvenient to use.<br>● 2-78:   User: The 3D effect of your software is very good |
| version release management | ● 2-105:   Designer: There will be an update. Submitting to the app store / releasing the Android version next Monday, and releasing the Windows version over the weekend.<br>● 2-114:   Designer: The Apple Store is still reviewing, let's upload the PC version first |

## 4.2 Chunk-level LLM Coding

| Label | Examples |
|-------|----------|
| apologies and politeness | ● 2-50:    Designer: Sorry for the late-night disturbance :) |
| collaborative problem-solving | ● 2-77:    Designer: Hope to figure out electromagnetism before the end of the year<br>● 2-82:    User: Don't aim for completeness, it should be categorized and refined one by one<br>● 2-71:    Designer: Actually, we are researching the bidirectional conversion between physical diagrams and circuit diagrams |
| communication of updates | ● 2-105:  Designer: There will be an update. Submitting to the app store / releasing the Android version next Monday, and releasing the Windows version over the weekend.<br>● 2-122:  Designer: The one in the group is the PC version<br>● 2-117:  Designer: The new version has already been sent in the group |
| community engagement | ● 2-2:     User: PC version?<br>● 2-8:     User: First, let's pay homage to the experts, then I'll make a small request. Could you create a dynamic demonstration of mechanical waves and mechanical vibrations, such as the propagation of transverse and longitudinal waves, wave superposition, diffraction, and interference? Also, for optical experiments, it would be great to have optical benches, single slits, double slits, and polarizers to demonstrate optical experiments.<br>● 2-12:   User: [Emoji] Hello everyone, may I ask where I can download the PC / interactive whiteboard version of Physics Lab?<br>● 2-11:   User: Okay, okay~ Thank you for your hard work.<br>● 2-58:   User: Hello everyone<br>● 2-59:   Designer: Hello :)<br>● 2-80:   User: It's already starting to take shape [Emoji]<br>● 2-103:  Designer: Thank you all for your support. We will do better!<br>● 2-104:  User: Will there be an update this week? [Emoji]<br>● 2-105:  Designer: There will be an update. Submitting to the app store / releasing the Android version next Monday, and releasing the Windows version over the weekend. |
| community feedback | ● 2-70:   User: This is good, I hope it can be very convenient to draw circuit diagrams, and I also hope there is a function to hide the background grid with one click, which is convenient for us teachers to take screenshots for test papers. Currently, there is no software that makes it easy to draw circuit diagrams for exam questions<br>● 2-73:   User: Can you also include mechanics experiments?<br>● 2-78:   User: The 3D effect of your software is very good |

| | | |
|---|---|---|
| community feedback loop | • 2-26: | User: This software is great! It's quite practical, unlike some software that tries to be comprehensive but ends up being inconvenient to use. |
| | • 2-31: | User: It is recommended to add an exit button function to the PC version. Many regions now use all-in-one touch screen machines without physical keyboards. |
| | • 2-38: | User: If there could be an export function, or the ability to save or import experiments, it would be convenient. We could set up the parameters in the office and directly import them in class. |
| community support | • 2-103: | Designer: Thank you all for your support. We will do better! |
| | • 2-112: | User: Yes, the spirit of craftsmanship that strives for perfection 🤔 |
| | • 2-107: | User: [Emoji] |
| community updates | • 2-15: | Designer: From the next update, dry batteries will no longer be ideal components (you can remove the internal resistance to simulate). |
| | • 2-16: | Designer: The student power supply supports both DC and AC and is an ideal component. |
| | • 2-29: | Designer: Hello everyone~ Version 1.01 has been released in the group files, the updates are as follows: |
| context of use | • 2-41: | User: Generally not. Ever since an adult image popped up during a major city-level open class, the school has disabled the network on classroom computers [Emoji]. |
| | • 2-31: | User: It is recommended to add an exit button function to the PC version. Many regions now use all-in-one touch screen machines without physical keyboards. |
| | • 2-38: | User: If there could be an export function, or the ability to save or import experiments, it would be convenient. We could set up the parameters in the office and directly import them in class. |
| contextual constraints | • 2-40: | Designer: Does the class have internet? |
| | • 2-41: | User: Generally not. Ever since an adult image popped up during a major city-level open class, the school has disabled the network on classroom computers [Emoji]. |
| | • 2-44: | User: Most schools will disable the network on classroom computers to prevent students from going online or to avoid various software auto-downloads from slowing down the computers. |
| design challenges | • 2-88: | Designer: For electricity, we plan to create a function for simulated experiment assessment; it will add many experimental details, such as zero adjustment (of course, only useful in specific modes, otherwise it would be cumbersome to use regularly) |
| | • 2-101: | Designer: This version of the circuit diagram is for testing only... You can drag the Editor because the auto-layout algorithm is not very stable and may produce some odd results |
| | • 2-98: | Designer: Make it simpler... |

| design philosophy | • 2-66: | Designer: We try to design it so that it can be used without additional instructions, or we will provide some prompts when you open it for the first time based on feedback |
| | • 2-84: | Designer: Physics mainly focuses on electricity and mechanics; other directions don't have much room for free experiments before high school. In the future, we might do some demonstration experiments based on everyone's needs, but it won't be as open as electricity |
| | • 2-74: | Designer: For example, you can see the corresponding circuit diagram after connecting the physical diagram, or vice versa |
| designer responsiveness | • 2-3: | Designer: I'll upload one now... Are you a teacher? |
| | • 2-6: | Designer: Hello everyone~ The development plan and PC download address are in the group announcement. |
| | • 2-9: | Designer: Hmm... We will probably prioritize completing the electrical section first, then magnetism, and then other parts. |
| | • 2-17: | Designer: Okay. Please give more suggestions! |
| | • 2-25: | Designer: No need to be polite, if you encounter any problems during use, you can directly mention them in the group. |
| | • 2-27: | Designer: XP compatibility might need to be looked into later... probably need to install a virtual machine. |
| | • 2-32: | Designer: Hmm... I'll add it in the next update. PC updates are usually the fastest, so there will probably be another round this weekend. |
| | • 2-39: | Designer: There will be. |
| | • 2-43: | Designer: It seems that local export needs to be supported. |
| | • 2-45: | Designer: Hmm hmm. |
| development prioritization | • 2-9: | Designer: Hmm... We will probably prioritize completing the electrical section first, then magnetism, and then other parts. |
| | • 2-10: | Designer: Before starting mechanics, we will gather opinions again~ Otherwise, I'm afraid I won't remember everything. |
| | • 2-15: | Designer: From the next update, dry batteries will no longer be ideal components (you can remove the internal resistance to simulate). |
| development updates | • 2-49: | Designer: Version 1.0.1 of Physics Lab. It is expected to take another one or two days in the Apple market (waiting for review). |
| | • 2-55: | Designer: We will also update the multimeter and powered solenoid this week. If there are any other features, characteristics, or electronic components you hope to include in the production plan, please leave a message in the group, and we will consider it as much as possible. The previously mentioned feature of saving experiments to the cloud/local is also in the plan. Thank you, everyone! |
| educational focus | • 2-84: | Designer: Physics mainly focuses on electricity and mechanics; other directions don't have much room for free experiments before high school. In the future, we might do some demonstration experiments based on everyone's needs, but it won't be as open as electricity |
| | • 2-79: | Designer: With bidirectional conversion, you can directly do problems in the application |
| | • 2-67: | User: Mainly, the school is building an information-based school |
| encouragement of user feedback | • 2-17: | Designer: Okay. Please give more suggestions! |
| expression of gratitude | • 2-11: | User: Okay, okay~ Thank you for your hard work. |
| | • 2-13: | User: I saw the group files, thank you. |
| feature announcements | | |
| feature request | • 2-8: | User: First, let's pay homage to the experts, then I'll make a small request. Could you create a dynamic demonstration of mechanical waves and mechanical vibrations, such as the propagation of transverse and longitudinal waves, wave superposition, diffraction, and interference? Also, for optical experiments, it would be great to have optical benches, single slits, double slits, and polarizers to demonstrate optical experiments. |
| | • 2-12: | User: [Emoji] Hello everyone, may I ask where I can download the PC / interactive whiteboard version of Physics Lab? |
| feature updates | • 2-55: | Designer: We will also update the multimeter and powered solenoid this week. If there are any other features, characteristics, or electronic components you hope to include in the production plan, please leave a message in the group, and we will consider it as much as possible. The previously mentioned feature of saving experiments to the cloud/local is also in the plan. Thank you, everyone! |
| | • 2-63: | Designer: There will be: multimeter; powered solenoid; semiconductor capacitor; support for conversion to ideal ammeter (more convenient for problem-solving and middle school teaching) |
| | • 2-88: | Designer: For electricity, we plan to create a function for simulated experiment assessment; it will add many experimental details, such as zero adjustment (of course, only useful in specific modes, otherwise it would be cumbersome to use regularly) |
| future plans | • 2-75: | Designer: Mechanics will have to wait until electromagnetism is figured out; it will take some more time |
| | • 2-87: | Designer: We'll see if there's a suitable opportunity to do one later |
| | • 2-77: | Designer: Hope to figure out electromagnetism before the end of the year |

| gratitude and acknowledgment | • 2-8: | User: First, let's pay homage to the experts, then I'll make a small request. Could you create a dynamic demonstration of mechanical waves and mechanical vibrations, such as the propagation of transverse and longitudinal waves, wave superposition, diffraction, and interference? Also, for optical experiments, it would be great to have optical benches, single slits, double slits, and polarizers to demonstrate optical experiments. |
| | • 2-11: | User: Okay, okay~ Thank you for your hard work. |
| | • 2-13: | User: I saw the group files, thank you. |
| informal communication | • 2-1: | Designer: @Morning Tea Moonlight How can I upload a high-definition, uncensored version of this crappy avatar? |
| | • 2-3: | Designer: I'll upload one now... Are you a teacher? |
| | • 2-42: | Designer: 😂 |
| | • 2-47: | Designer: Haha. Are you a middle school or high school teacher? |
| | • 2-50: | Designer: Sorry for the late-night disturbance :) |
| iterative development | • 2-27: | Designer: XP compatibility might need to be looked into later... probably need to install a virtual machine. |
| | • 2-29: | Designer: Hello everyone~ Version 1.01 has been released in the group files, the updates are as follows: |
| | • 2-35: | Designer: But you can't bring it up without touching the input, that's probably the case. |
| | • 2-95: | Designer: We are working on this part |
| | • 2-101: | Designer: This version of the circuit diagram is for testing only... You can drag the Editor because the auto-layout algorithm is not very stable and may produce some odd results |
| | • 2-106: | Designer: The update is quite large... |
| non-verbal communication | • 2-56: | User: [Emoji] |
| | • 2-57: | User: [Emoji][Emoji] |
| participatory design | • 2-7: | Designer: If you have any suggestions or requirements, feel free to bring them up. |
| | • 2-8: | User: First, let's pay homage to the experts, then I'll make a small request. Could you create a dynamic demonstration of mechanical waves and mechanical vibrations, such as the propagation of transverse and longitudinal waves, wave superposition, diffraction, and interference? Also, for optical experiments, it would be great to have optical benches, single slits, double slits, and polarizers to demonstrate optical experiments. |
| | • 2-10: | Designer: Before starting mechanics, we will gather opinions again~ Otherwise, I'm afraid I won't remember everything. |
| | • 2-16: | Designer: The student power supply supports both DC and AC and is an ideal component. |
| | • 2-17: | Designer: Okay. Please give more suggestions! |
| | • 2-25: | Designer: No need to be polite, if you encounter any problems during use, you can directly mention them in the group. |

| participatory design | • 2-27: | Designer: XP compatibility might need to be looked into later... probably need to install a virtual machine. |
|---|---|---|
| | • 2-31: | User: It is recommended to add an exit button function to the PC version. Many regions now use all-in-one touch screen machines without physical keyboards. |
| | • 2-33: | Designer: Speaking of which, doesn't that mean every place where numbers are input should have a soft keyboard? |
| | • 2-38: | User: If there could be an export function, or the ability to save or import experiments, it would be convenient. We could set up the parameters in the office and directly import them in class. |
| | • 2-43: | Designer: It seems that local export needs to be supported. |
| | • 2-53: | Designer: Group sharing. |
| | • 2-55: | Designer: We will also update the multimeter and powered solenoid this week. If there are any other features, characteristics, or electronic components you hope to include in the production plan, please leave a message in the group, and we will consider it as much as possible. The previously mentioned feature of saving experiments to the cloud/local is also in the plan. Thank you, everyone! |
| | • 2-90: | Designer: When we are ready to start, we will consult everyone in the group |
| | • 2-91: | Designer: Consulting the teachers in the group: which type of intersection is used in the circuit diagrams in the current textbooks? [Image] |
| | • 2-100: | Designer: So, use both? |
| | • 2-101: | Designer: This version of the circuit diagram is for testing only... You can drag the Editor because the auto-layout algorithm is not very stable and may produce some odd results |
| | • 2-109: | Designer: Don't worry, don't worry, it will come, just optimizing the circuit diagram one last time. |
| | • 2-110: | Designer: Although you can manually adjust the layout |
| product update information | • 2-15: | Designer: From the next update, dry batteries will no longer be ideal components (you can remove the internal resistance to simulate). |
| | • 2-16: | Designer: The student power supply supports both DC and AC and is an ideal component. |
| | • 2-20: | Designer: It does not support the XP system. |
| recognition of effort | • 2-8: | User: First, let's pay homage to the experts, then I'll make a small request. Could you create a dynamic demonstration of mechanical waves and mechanical vibrations, such as the propagation of transverse and longitudinal waves, wave superposition, diffraction, and interference? Also, for optical experiments, it would be great to have optical benches, single slits, double slits, and polarizers to demonstrate optical experiments. |
| | • 2-11: | User: Okay, okay~ Thank you for your hard work. |
| | • 2-13: | User: I saw the group files, thank you. |
| request for information | ● 2-12: | User: [Emoji] Hello everyone, may I ask where I can download the PC / interactive whiteboard version of Physics Lab? |
| | ● 2-18: | User: How do you use the PC version? |
| resource sharing | ● 2-6: | Designer: Hello everyone~ The development plan and PC download address are in the group announcement. |
| | ● 2-3: | Designer: I'll upload one now... Are you a teacher? |
| | ● 2-13: | User: I saw the group files, thank you. |
| | ● 2-53: | Designer: Group sharing. |
| | ● 2-54: | Designer: You can download Windows in the group files. |
| role identification | ● 2-3: | Designer: I'll upload one now... Are you a teacher? |
| | ● 2-4: | User: Yes. |
| sharing resources | ● 2-13: | User: I saw the group files, thank you. |
| | ● 2-14: | Designer: [Image] |
| technical constraints | ● 2-27: | Designer: XP compatibility might need to be looked into later... probably need to install a virtual machine. |
| | ● 2-28: | Designer: Theoretically, it should be compatible (but the machine itself shouldn't be too old, probably from 2008 onwards). |
| | ● 2-34: | Designer: Touch screens have their own on-screen keyboards. |
| technical inquiry | ● 2-2: | User: PC version? |
| technical limitations | ● 2-20: | Designer: It does not support the XP system. |
| technical support | ● 2-1: | Designer: @Morning Tea Moonlight How can I upload a high-definition, uncensored version of this crappy avatar? |
| | ● 2-3: | Designer: I'll upload one now... Are you a teacher? |
| | ● 2-12: | User: [Emoji] Hello everyone, may I ask where I can download the PC / interactive whiteboard version of Physics Lab? |
| | ● 2-20: | Designer: It does not support the XP system. |
| | ● 2-23: | Designer: 7 is okay. |
| | ● 2-28: | Designer: Theoretically, it should be compatible (but the machine itself shouldn't be too old, probably from 2008 onwards). |
| technical updates | ● 2-15: | Designer: From the next update, dry batteries will no longer be ideal components (you can remove the internal resistance to simulate). |
| | ● 2-16: | Designer: The student power supply supports both DC and AC and is an ideal component. |
| | ● 2-14: | Designer: [Image] |
| transparency in development | ● 2-114: | Designer: The Apple Store is still reviewing, let's upload the PC version first |
| | ● 2-103: | Designer: Thank you all for your support. We will do better! |
| | ● 2-113: | Designer: This belongs to the kind of feature that, once done, will ensure long-term stability... Adding various components is actually simpler |

| user background | • 2-48: | User: I taught high school for seven years, and now I've been teaching middle school for eight years. |
|---|---|---|
| user engagement | • 2-1: | Designer: @Morning Tea Moonlight How can I upload a high-definition, uncensored version of this crappy avatar? |
| | • 2-2: | User: PC version? |
| | • 2-4: | User: Yes. |
| | • 2-104: | User: Will there be an update this week? [Emoji] |
| | • 2-112: | User: Yes, the spirit of craftsmanship that strives for perfection 🤔 |
| | • 2-125: | User: Has the Android version not been updated yet? |
| user experience | • 2-22: | User: Win7 should be fine. |
| | • 2-24: | User: Thank you. |
| | • 2-26: | User: This software is great! It's quite practical, unlike some software that tries to be comprehensive but ends up being inconvenient to use. |
| user experience and impact | • 2-46: | User: The simulation effect of this software is really good. When I used it in class yesterday, the students were amazed. |
| | • 2-48: | User: I taught high school for seven years, and now I've been teaching middle school for eight years. |
| | • 2-50: | Designer: Sorry for the late-night disturbance :) |
| user feedback | • 2-89: | User: This idea is really good |
| | • 2-93: | User: Yes, the common one is still the old style |
| | • 2-99: | User: But it's better to use the right side for non-crossing |
| user feedback and suggestions | • 2-18: | User: How do you use the PC version? |
| | • 2-21: | User: Dizzy, the computer system for teachers at our school is quite old. |
| | • 2-26: | User: This software is great! It's quite practical, unlike some software that tries to be comprehensive but ends up being inconvenient to use. |
| | • 2-38: | User: If there could be an export function, or the ability to save or import experiments, it would be convenient. We could set up the parameters in the office and directly import them in class. |
| | • 2-44: | User: Most schools will disable the network on classroom computers to prevent students from going online or to avoid various software auto-downloads from slowing down the computers. |
| | • 2-46: | User: The simulation effect of this software is really good. When I used it in class yesterday, the students were amazed. |
| user inquiries | • 2-51: | User: How to download? |
| | • 2-52: | User: Excuse me. |
| user support | • 2-65: | Designer: What problems did you encounter during use? |
| | • 2-69: | Designer: Oh, please message me privately and tell me what you roughly need |
| | • 2-66: | Designer: We try to design it so that it can be used without additional instructions, or we will provide some prompts when you open it for the first time based on feedback |

## 4.3 Item-level LLM Coding

| Label | Examples |
|---|---|
| acknowledgment | <ul><li>2-11:  User: Okay, okay~ Thank you for your hard work.</li><li>2-13:  User: I saw the group files, thank you.</li><li>2-45:  Designer: Hmm hmm.</li><li>2-83:  Designer: Hmm.</li><li>2-97:  Designer: Okay.</li></ul> |
| agreement | <ul><li>2-85:  User: Indeed</li></ul> |
| apology | <ul><li>2-50:  Designer: Sorry for the late-night disturbance :)</li></ul> |
| augmented reality | <ul><li>2-115:  Designer: v1.0.2 - 17/11/29<br># New Features<br>Multi-language support: Japanese, German, French.<br>Preliminary support for converting circuit components to circuit diagrams.<br>Support for augmented reality mode on the latest iOS devices.<br>More realistic 3D appearance for various meters.<br><br># Update Preview<br>The next version will provide multimeter and meter zeroing.<br>The next version will provide AC waveform display functionality.<br>The next version will provide capacitors and energized solenoids.<br><br># Component Property Adjustments<br>Dry cells can now be ideal power sources.<br>All meters can now switch between ideal/realistic modes.<br><br># Special Tips<br>The Esc shortcut key will now minimize the window.<br>The Ctrl+E shortcut key can switch between exam and normal modes.</li></ul> |
| auto layout optimization | <ul><li>2-111:  Designer: It's still better to make the auto-generated one as good as possible</li></ul> |
| avatar customization | <ul><li>2-1:  Designer: @Morning Tea Moonlight How can I upload a high-definition, uncensored version of this crappy avatar?</li></ul> |

| bug fixes | • 2-30: | Designer: New Features<br>Electronic components will be damaged after a short process, rather than immediately.<br>Clearing the desktop will now display a confirmation interface.<br>Supports undoing the creation and deletion of wires and components.<br>Appliances now display the effective value of alternating current.<br>(PC) You can now exit the application using the Esc key.<br><br>New Components<br>Added a sensitive ammeter.<br>Added a student power supply (ideal AC/DC power supply).<br><br>Adjustments to Component Properties<br>Batteries now have adjustable internal resistance and are no longer ideal power sources.<br>Incandescent bulbs now have volt-ampere characteristic parameters and are no longer ideal resistors.<br>The resistance law experimenter now uses real formulas for calculations, with adjustable parameters.<br>Hidden terminal blocks 3 and 4.<br><br>Bug Fixes<br>Fixed an issue with unit conversion in Editor properties.<br>Terminal arrows no longer show jumping animations.<br>There may have been calculation errors with certain circuit connections. |
|---|---|---|
| clarification request | • 2-94: | Designer: Uh... left side or right side |
| classroom application | • 2-38:<br><br><br><br>• 2-46: | User: If there could be an export function, or the ability to save or import experiments, it would be convenient. We could set up the parameters in the office and directly import them in class.<br>User: The simulation effect of this software is really good. When I used it in class yesterday, the students were amazed. |
| classroom environment | • 2-41:<br><br><br>• 2-44: | User: Generally not. Ever since an adult image popped up during a major city-level open class, the school has disabled the network on classroom computers [Emoji].<br>User: Most schools will disable the network on classroom computers to prevent students from going online or to avoid various software auto-downloads from slowing down the computers. |
| classroom setup | • 2-40: | Designer: Does the class have internet? |
| collaborative learning | ● 2-53: | Designer: Group sharing. |
| communication tone | ● 2-1: | Designer: @Morning Tea Moonlight How can I upload a high-definition, uncensored version of this crappy avatar? |
| community building | ● 2-24: | User: Thank you. |
| | ● 2-42: | Designer: 😂 |
| community communication | ● 2-29: | Designer: Hello everyone~ Version 1.01 has been released in the group files, the updates are as follows: |
| community context | ● 2-67: | User: Mainly, the school is building an information-based school |
| community engagement | ● 2-8: | User: First, let's pay homage to the experts, then I'll make a small request. Could you create a dynamic demonstration of mechanical waves and mechanical vibrations, such as the propagation of transverse and longitudinal waves, wave superposition, diffraction, and interference? Also, for optical experiments, it would be great to have optical benches, single slits, double slits, and polarizers to demonstrate optical experiments. |
| | ● 2-17: | Designer: Okay. Please give more suggestions! |
| | ● 2-25: | Designer: No need to be polite, if you encounter any problems during use, you can directly mention them in the group. |
| | ● 2-37: | User: @Designer Yes, yes. |
| | ● 2-61: | Designer: [Image] Updates in preparation |
| community interaction | ● 2-52: | User: Excuse me. |
| | ● 2-57: | User: [Emoji][Emoji] |
| | ● 2-58: | User: Hello everyone |
| | ● 2-59: | Designer: Hello :) |
| | ● 2-60: | User: Are there any other updates recently? |
| | ● 2-65: | Designer: What problems did you encounter during use? |
| | ● 2-69: | Designer: Oh, please message me privately and tell me what you roughly need |
| | ● 2-76: | User: Oh, this is nice |
| | ● 2-78: | User: The 3D effect of your software is very good |
| | ● 2-80: | User: It's already starting to take shape [Emoji] |
| | ● 2-83: | Designer: Hmm. |
| | ● 2-85: | User: Indeed |
| | ● 2-86: | Designer: I myself have some interest in chemistry |
| | ● 2-87: | Designer: We'll see if there's a suitable opportunity to do one later |
| | ● 2-89: | User: This idea is really good |
| | ● 2-93: | User: Yes, the common one is still the old style |
| | ● 2-96: | User: Left side |
| | ● 2-99: | User: But it's better to use the right side for non-crossing |

| | |
|---|---|
| community interaction | • 2-104: User: Will there be an update this week? [Emoji]<br>• 2-112: User: Yes, the spirit of craftsmanship that strives for perfection 🤔<br>• 2-120: User: ok<br>• 2-125: User: Has the Android version not been updated yet?<br>• 2-126: User: Updated |
| community involvement | • 2-55: Designer: We will also update the multimeter and powered solenoid this week. If there are any other features, characteristics, or electronic components you hope to include in the production plan, please leave a message in the group, and we will consider it as much as possible. The previously mentioned feature of saving experiments to the cloud/local is also in the plan. Thank you, everyone!<br>• 2-90: Designer: When we are ready to start, we will consult everyone in the group<br>• 2-91: Designer: Consulting the teachers in the group: which type of intersection is used in the circuit diagrams in the current textbooks? [Image]<br>• 2-92: Designer: [Image] This one, right? |
| community member identification | • 2-4: User: Yes. |
| community norms | • 2-50: Designer: Sorry for the late-night disturbance :) |
| community resource | • 2-53: Designer: Group sharing.<br>• 2-64: User: Is there a user manual?<br>• 2-68: User: Need the user manual for this software |
| community support | • 2-11: User: Okay, okay~ Thank you for your hard work.<br>• 2-54: Designer: You can download Windows in the group files.<br>• 2-63: Designer: There will be: multimeter; powered solenoid; semiconductor capacitor; support for conversion to ideal ammeter (more convenient for problem-solving and middle school teaching)<br>• 2-70: User: This is good, I hope it can be very convenient to draw circuit diagrams, and I also hope there is a function to hide the background grid with one click, which is convenient for us teachers to take screenshots for test papers. Currently, there is no software that makes it easy to draw circuit diagrams for exam questions<br>• 2-103: Designer: Thank you all for your support. We will do better! |
| community update | • 2-6: Designer: Hello everyone~ The development plan and PC download address are in the group announcement.<br>• 2-62: Designer: This is quite complex, so it will take more time... Hopefully, it can be released this week |

| comparative feedback | • 2-26: | User: This software is great! It's quite practical, unlike some software that tries to be comprehensive but ends up being inconvenient to use. |
|---|---|---|
| compatibility assurance | • 2-28: | Designer: Theoretically, it should be compatible (but the machine itself shouldn't be too old, probably from 2008 onwards). |
| component adjustments | • 2-115: | Designer: v1.0.2 - 17/11/29<br># New Features<br>Multi-language support: Japanese, German, French.<br>Preliminary support for converting circuit components to circuit diagrams.<br>Support for augmented reality mode on the latest iOS devices.<br>More realistic 3D appearance for various meters.<br><br># Update Preview<br>The next version will provide multimeter and meter zeroing.<br>The next version will provide AC waveform display functionality.<br>The next version will provide capacitors and energized solenoids.<br><br># Component Property Adjustments<br>Dry cells can now be ideal power sources.<br>All meters can now switch between ideal/realistic modes.<br><br># Special Tips<br>The Esc shortcut key will now minimize the window.<br>The Ctrl+E shortcut key can switch between exam and normal modes. |
| component functionality | • 2-16: | Designer: The student power supply supports both DC and AC and is an ideal component. |
| component integration | • 2-113: | Designer: This belongs to the kind of feature that, once done, will ensure long-term stability... Adding various components is actually simpler |
| component update | • 2-15: | Designer: From the next update, dry batteries will no longer be ideal components (you can remove the internal resistance to simulate). |
| consultation | • 2-90:<br><br>• 2-91: | Designer: When we are ready to start, we will consult everyone in the group<br>Designer: Consulting the teachers in the group: which type of intersection is used in the circuit diagrams in the current textbooks? [Image] |
| craftsmanship | ● 2-112: | User: Yes, the spirit of craftsmanship that strives for perfection 🤔 |
| cross platform usage | ● 2-2: | User: PC version? |
| design adaptation | ● 2-43: | Designer: It seems that local export needs to be supported. |
| design decision | ● 2-100: | Designer: So, use both? |
| design philosophy | ● 2-66:<br><br><br>● 2-82: | Designer: We try to design it so that it can be used without additional instructions, or we will provide some prompts when you open it for the first time based on feedback<br>User: Don't aim for completeness, it should be categorized and refined one by one |
| design refinement | ● 2-98: | Designer: Make it simpler... |
| designer acknowledgment | ● 2-32: | Designer: Hmm... I'll add it in the next update. PC updates are usually the fastest, so there will probably be another round this weekend. |
| designer confirmation | ● 2-39: | Designer: There will be. |
| designer engagement | ● 2-45:<br>● 2-86: | Designer: Hmm hmm.<br>Designer: I myself have some interest in chemistry |
| designer response | ● 2-59:<br>● 2-83: | Designer: Hello :)<br>Designer: Hmm. |
| designer user interaction | ● 2-3: | Designer: I'll upload one now... Are you a teacher? |
| development plan | ● 2-6: | Designer: Hello everyone~ The development plan and PC download address are in the group announcement. |
| development prioritization | ● 2-9: | Designer: Hmm... We will probably prioritize completing the electrical section first, then magnetism, and then other parts. |
| development timeline | ● 2-49:<br><br><br>● 2-62:<br><br>● 2-75:<br><br>● 2-77: | Designer: Version 1.0.1 of Physics Lab. It is expected to take another one or two days in the Apple market (waiting for review).<br>Designer: This is quite complex, so it will take more time... Hopefully, it can be released this week<br>Designer: Mechanics will have to wait until electromagnetism is figured out; it will take some more time<br>Designer: Hope to figure out electromagnetism before the end of the year |

| | |
|---|---|
| development transparency | • 2-55:   Designer: We will also update the multimeter and powered solenoid this week. If there are any other features, characteristics, or electronic components you hope to include in the production plan, please leave a message in the group, and we will consider it as much as possible. The previously mentioned feature of saving experiments to the cloud/local is also in the plan. Thank you, everyone!<br>• 2-61:   Designer: [Image] Updates in preparation |
| download inquiry | • 2-12:   User: [Emoji] Hello everyone, may I ask where I can download the PC / interactive whiteboard version of Physics Lab?<br>• 2-51:   User: How to download? |
| download instructions | • 2-54:   Designer: You can download Windows in the group files. |
| educational context | • 2-21:   User: Dizzy, the computer system for teachers at our school is quite old.<br>• 2-36:   Designer: Last time I tried it in middle school, there's a trick where you click the link button in the top right, and when the browser pops up, it's no longer full screen... = =<br>• 2-67:   User: Mainly, the school is building an information-based school |
| educational tools | • 2-8:   User: First, let's pay homage to the experts, then I'll make a small request. Could you create a dynamic demonstration of mechanical waves and mechanical vibrations, such as the propagation of transverse and longitudinal waves, wave superposition, diffraction, and interference? Also, for optical experiments, it would be great to have optical benches, single slits, double slits, and polarizers to demonstrate optical experiments.<br>• 2-16:   Designer: The student power supply supports both DC and AC and is an ideal component.<br>• 2-63:   Designer: There will be: multimeter; powered solenoid; semiconductor capacitor; support for conversion to ideal ammeter (more convenient for problem-solving and middle school teaching)<br>• 2-64:   User: Is there a user manual?<br>• 2-68:   User: Need the user manual for this software<br>• 2-70:   User: This is good, I hope it can be very convenient to draw circuit diagrams, and I also hope there is a function to hide the background grid with one click, which is convenient for us teachers to take screenshots for test papers. Currently, there is no software that makes it easy to draw circuit diagrams for exam questions<br>• 2-71:   Designer: Actually, we are researching the bidirectional conversion between physical diagrams and circuit diagrams |

| | | |
|---|---|---|
| educational tools | • 2-73: | User: Can you also include mechanics experiments? |
| | • 2-74: | Designer: For example, you can see the corresponding circuit diagram after connecting the physical diagram, or vice versa |
| | • 2-79: | Designer: With bidirectional conversion, you can directly do problems in the application |
| | • 2-81: | Designer: For example, see the circuit diagram to connect the physical diagram, or vice versa |
| | • 2-84: | Designer: Physics mainly focuses on electricity and mechanics; other directions don't have much room for free experiments before high school. In the future, we might do some demonstration experiments based on everyone's needs, but it won't be as open as electricity |
| | • 2-88: | Designer: For electricity, we plan to create a function for simulated experiment assessment; it will add many experimental details, such as zero adjustment (of course, only useful in specific modes, otherwise it would be cumbersome to use regularly) |
| | • 2-90: | Designer: When we are ready to start, we will consult everyone in the group |
| | • 2-91: | Designer: Consulting the teachers in the group: which type of intersection is used in the circuit diagrams in the current textbooks? [Image] |
| emoji | • 2-56: | User: [Emoji] |
| | • 2-107: | User: [Emoji] |
| | • 2-108: | User: [Emoji] |
| | • 2-116: | User: [Emoji] |
| | • 2-119: | User: [Emoji] |
| | • 2-121: | User: [Emoji] |
| emoji use | • 2-12: | User: [Emoji] Hello everyone, may I ask where I can download the PC / interactive whiteboard version of Physics Lab? |
| | • 2-57: | User: [Emoji][Emoji] |
| expert recognition | • 2-8: | User: First, let's pay homage to the experts, then I'll make a small request. Could you create a dynamic demonstration of mechanical waves and mechanical vibrations, such as the propagation of transverse and longitudinal waves, wave superposition, diffraction, and interference? Also, for optical experiments, it would be great to have optical benches, single slits, double slits, and polarizers to demonstrate optical experiments. |
| feature adjustment | • 2-43: | Designer: It seems that local export needs to be supported. |
| feature appreciation | • 2-76: | User: Oh, this is nice |
| | • 2-80: | User: It's already starting to take shape [Emoji] |
| | • 2-89: | User: This idea is really good |

| feature complexity | • 2-62: | Designer: This is quite complex, so it will take more time... Hopefully, it can be released this week |
|---|---|---|
| feature development | • 2-71: | Designer: Actually, we are researching the bidirectional conversion between physical diagrams and circuit diagrams |
| | • 2-72: | Designer: Of course, manual editing will also be allowed, but it might be a bit later |
| | • 2-88: | Designer: For electricity, we plan to create a function for simulated experiment assessment; it will add many experimental details, such as zero adjustment (of course, only useful in specific modes, otherwise it would be cumbersome to use regularly) |
| | • 2-95: | Designer: We are working on this part |
| | • 2-101: | Designer: This version of the circuit diagram is for testing only... You can drag the Editor because the auto-layout algorithm is not very stable and may produce some odd results |
| feature discussion | • 2-33: | Designer: Speaking of which, doesn't that mean every place where numbers are input should have a soft keyboard? |
| feature expansion | • 2-106: | Designer: The update is quite large... |
| feature explanation | • 2-74: | Designer: For example, you can see the corresponding circuit diagram after connecting the physical diagram, or vice versa |
| | • 2-79: | Designer: With bidirectional conversion, you can directly do problems in the application |
| | • 2-81: | Designer: For example, see the circuit diagram to connect the physical diagram, or vice versa |
| feature flexibility | • 2-110: | Designer: Although you can manually adjust the layout |
| feature guidance | • 2-124: | Designer: After connecting the student power supply, you need to turn on the switch |
| feature implementation | • 2-32: | Designer: Hmm... I'll add it in the next update. PC updates are usually the fastest, so there will probably be another round this weekend. |
| | • 2-39: | Designer: There will be. |
| feature improvement | • 2-99: | User: But it's better to use the right side for non-crossing |
| | • 2-111: | Designer: It's still better to make the auto-generated one as good as possible |
| feature list | • 2-63: | Designer: There will be: multimeter; powered solenoid; semiconductor capacitor; support for conversion to ideal ammeter (more convenient for problem-solving and middle school teaching) |
| feature optimization | • 2-109: | Designer: Don't worry, don't worry, it will come, just optimizing the circuit diagram one last time. |

| feature preview | ● 2-61: Designer: [Image] Updates in preparation |
|---|---|
| feature prioritization | ● * 2-75: Designer: Mechanics will have to wait until electromagnetism is figured out; it will take some more time<br>● * 2-77: Designer: Hope to figure out electromagnetism before the end of the year<br>● * 2-84: Designer: Physics mainly focuses on electricity and mechanics; other directions don't have much room for free experiments before high school. In the future, we might do some demonstration experiments based on everyone's needs, but it won't be as open as electricity |
| feature release | ● 2-29: Designer: Hello everyone~ Version 1.01 has been released in the group files, the updates are as follows: |
| feature request | ● 2-8: User: First, let's pay homage to the experts, then I'll make a small request. Could you create a dynamic demonstration of mechanical waves and mechanical vibrations, such as the propagation of transverse and longitudinal waves, wave superposition, diffraction, and interference? Also, for optical experiments, it would be great to have optical benches, single slits, double slits, and polarizers to demonstrate optical experiments.<br>● 2-31: User: It is recommended to add an exit button function to the PC version. Many regions now use all-in-one touch screen machines without physical keyboards.<br>● 2-38: User: If there could be an export function, or the ability to save or import experiments, it would be convenient. We could set up the parameters in the office and directly import them in class.<br>● 2-70: User: This is good, I hope it can be very convenient to draw circuit diagrams, and I also hope there is a function to hide the background grid with one click, which is convenient for us teachers to take screenshots for test papers. Currently, there is no software that makes it easy to draw circuit diagrams for exam questions<br>● 2-73: User: Can you also include mechanics experiments? |
| feature roadmap | ● 2-9: Designer: Hmm... We will probably prioritize completing the electrical section first, then magnetism, and then other parts. |
| feature simplification | ● 2-98: Designer: Make it simpler... |
| feature stability | ● 2-113: Designer: This belongs to the kind of feature that, once done, will ensure long-term stability... Adding various components is actually simpler |
| feature suggestion | ● 2-82: User: Don't aim for completeness, it should be categorized and refined one by one |

| feature update | • 2-30: | Designer: New Features<br>Electronic components will be damaged after a short process, rather than immediately.<br>Clearing the desktop will now display a confirmation interface.<br>Supports undoing the creation and deletion of wires and components.<br>Appliances now display the effective value of alternating current.<br>(PC) You can now exit the application using the Esc key.<br><br>New Components<br>Added a sensitive ammeter.<br>Added a student power supply (ideal AC/DC power supply).<br><br>Adjustments to Component Properties<br>Batteries now have adjustable internal resistance and are no longer ideal power sources.<br>Incandescent bulbs now have volt-ampere characteristic parameters and are no longer ideal resistors.<br>The resistance law experimenter now uses real formulas for calculations, with adjustable parameters.<br>Hidden terminal blocks 3 and 4.<br><br>Bug Fixes<br>Fixed an issue with unit conversion in Editor properties.<br>Terminal arrows no longer show jumping animations.<br>There may have been calculation errors with certain circuit connections. |
| | • 2-55: | Designer: We will also update the multimeter and powered solenoid this week. If there are any other features, characteristics, or electronic components you hope to include in the production plan, please leave a message in the group, and we will consider it as much as possible. The previously mentioned feature of saving experiments to the cloud/local is also in the plan. Thank you, everyone! |
| feedback solicitation | • 2-7: | Designer: If you have any suggestions or requirements, feel free to bring them up. |
| future planning | • 2-27: | Designer: XP compatibility might need to be looked into later... probably need to install a virtual machine. |
| | • 2-39: | Designer: There will be. |
| future plans | • 2-86: | Designer: I myself have some interest in chemistry |
| | • 2-87: | Designer: We'll see if there's a suitable opportunity to do one later |

| general announcement | • 2-6: | Designer: Hello everyone~ The development plan and PC download address are in the group announcement. |
|---|---|---|
| gratitude | • 2-103: | Designer: Thank you all for your support. We will do better! |
| gratitude expression | • 2-11: | User: Okay, okay~ Thank you for your hard work. |
| greeting | • 2-12: | User: [Emoji] Hello everyone, may I ask where I can download the PC / interactive whiteboard version of Physics Lab? |
| | • 2-58: | User: Hello everyone |
| | • 2-59: | Designer: Hello :) |
| group communication | • 2-117: | Designer: The new version has already been sent in the group |
| group sharing | • 2-53: | Designer: Group sharing. |
| high school | • 2-48: | User: I taught high school for seven years, and now I've been teaching middle school for eight years. |
| humor | • 2-42: | Designer: 😂 |
| image sharing | • 2-0: | Designer: [Image] |
| | • 2-14: | Designer: [Image] |
| | • 2-19: | User: [Image] |
| | • 2-102: | Designer: [Image] |
| import/export functionality | • 2-38: | User: If there could be an export function, or the ability to save or import experiments, it would be convenient. We could set up the parameters in the office and directly import them in class. |
| informal communication | • 2-42: | Designer: 😂 |
| | • 2-45: | Designer: Hmm hmm. |
| | • 2-47: | Designer: Haha. Are you a middle school or high school teacher? |
| | • 2-50: | Designer: Sorry for the late-night disturbance :) |
| information dissemination | • 2-53: | Designer: Group sharing. |
| information retrieval | • 2-13: | User: I saw the group files, thank you. |
| input methods | • 2-33: | Designer: Speaking of which, doesn't that mean every place where numbers are input should have a soft keyboard? |
| instructional design | • 2-66: | Designer: We try to design it so that it can be used without additional instructions, or we will provide some prompts when you open it for the first time based on feedback |
| internet availability inquiry | • 2-40: | Designer: Does the class have internet? |
| internet restriction | ● 2-41: | User: Generally not. Ever since an adult image popped up during a major city-level open class, the school has disabled the network on classroom computers [Emoji]. |
| | ● 2-44: | User: Most schools will disable the network on classroom computers to prevent students from going online or to avoid various software auto-downloads from slowing down the computers. |
| iterative development | ● 2-10: | Designer: Before starting mechanics, we will gather opinions again~ Otherwise, I'm afraid I won't remember everything. |
| | ● 2-17: | Designer: Okay. Please give more suggestions! |
| local export necessity | ● 2-43: | Designer: It seems that local export needs to be supported. |
| long term planning | ● 2-113: | Designer: This belongs to the kind of feature that, once done, will ensure long-term stability... Adding various components is actually simpler |
| manual adjustment | ● 2-110: | Designer: Although you can manually adjust the layout |
| manual editing | ● 2-72: | Designer: Of course, manual editing will also be allowed, but it might be a bit later |
| memory aid | ● 2-10: | Designer: Before starting mechanics, we will gather opinions again~ Otherwise, I'm afraid I won't remember everything. |
| middle school | ● 2-48: | User: I taught high school for seven years, and now I've been teaching middle school for eight years. |
| multi language support | ● 2-115: | Designer: v1.0.2 - 17/11/29<br># New Features<br>Multi-language support: Japanese, German, French.<br>Preliminary support for converting circuit components to circuit diagrams.<br>Support for augmented reality mode on the latest iOS devices.<br>More realistic 3D appearance for various meters.<br><br># Update Preview<br>The next version will provide multimeter and meter zeroing.<br>The next version will provide AC waveform display functionality.<br>The next version will provide capacitors and energized solenoids.<br><br># Component Property Adjustments<br>Dry cells can now be ideal power sources.<br>All meters can now switch between ideal/realistic modes.<br><br># Special Tips |
| multi language support | | The Esc shortcut key will now minimize the window.<br>The Ctrl+E shortcut key can switch between exam and normal modes. |
| network policy | • 2-44: | User: Most schools will disable the network on classroom computers to prevent students from going online or to avoid various software auto-downloads from slowing down the computers. |
| new components | • 2-30: | Designer: New Features<br>Electronic components will be damaged after a short process, rather than immediately.<br>Clearing the desktop will now display a confirmation interface.<br>Supports undoing the creation and deletion of wires and components.<br>Appliances now display the effective value of alternating current.<br>(PC) You can now exit the application using the Esc key.<br><br>New Components<br>Added a sensitive ammeter.<br>Added a student power supply (ideal AC/DC power supply).<br><br>Adjustments to Component Properties<br>Batteries now have adjustable internal resistance and are no longer ideal power sources.<br>Incandescent bulbs now have volt-ampere characteristic parameters and are no longer ideal resistors.<br>The resistance law experimenter now uses real formulas for calculations, with adjustable parameters.<br>Hidden terminal blocks 3 and 4.<br><br>Bug Fixes<br>Fixed an issue with unit conversion in Editor properties.<br>Terminal arrows no longer show jumping animations.<br>There may have been calculation errors with certain circuit connections. |
| new features | • 2-115: | Designer: v1.0.2 - 17/11/29<br># New Features<br>Multi-language support: Japanese, German, French.<br>Preliminary support for converting circuit components to circuit diagrams.<br>Support for augmented reality mode on the latest iOS devices.<br>More realistic 3D appearance for various meters.<br><br># Update Preview |

| | |
|---|---|
| new features | The next version will provide multimeter and meter zeroing.<br>The next version will provide AC waveform display functionality.<br>The next version will provide capacitors and energized solenoids.<br><br># Component Property Adjustments<br>Dry cells can now be ideal power sources.<br>All meters can now switch between ideal/realistic modes.<br><br># Special Tips<br>The Esc shortcut key will now minimize the window.<br>The Ctrl+E shortcut key can switch between exam and normal modes. |
| new user interaction | • 2-12:   User: [Emoji] Hello everyone, may I ask where I can download the PC / interactive whiteboard version of Physics Lab?<br>• 2-18:   User: How do you use the PC version? |
| non verbal communication | • 2-57:   User: [Emoji][Emoji] |
| open communication | • 2-7:   Designer: If you have any suggestions or requirements, feel free to bring them up.<br>• 2-25:   Designer: No need to be polite, if you encounter any problems during use, you can directly mention them in the group. |
| opinion gathering | • 2-10:   Designer: Before starting mechanics, we will gather opinions again~ Otherwise, I'm afraid I won't remember everything. |
| outdated technology | • 2-21:   User: Dizzy, the computer system for teachers at our school is quite old. |
| participatory design | • 2-7:   Designer: If you have any suggestions or requirements, feel free to bring them up.<br>• 2-17:   Designer: Okay. Please give more suggestions!<br>• 2-55:   Designer: We will also update the multimeter and powered solenoid this week. If there are any other features, characteristics, or electronic components you hope to include in the production plan, please leave a message in the group, and we will consider it as much as possible. The previously mentioned feature of saving experiments to the cloud/local is also in the plan. Thank you, everyone!<br>• 2-66:   Designer: We try to design it so that it can be used without additional instructions, or we will provide some prompts when you open it for the first time based on feedback<br>• 2-70:   User: This is good, I hope it can be very convenient to draw circuit diagrams, and I also hope there is a function to hide the background grid with one click, which is convenient for us teachers to take screenshots for test papers. Currently, there |

| participatory design | | is no software that makes it easy to draw circuit diagrams for exam questions |
|---|---|---|
| | ● 2-71: | Designer: Actually, we are researching the bidirectional conversion between physical diagrams and circuit diagrams |
| | ● 2-72: | Designer: Of course, manual editing will also be allowed, but it might be a bit later |
| | ● 2-73: | User: Can you also include mechanics experiments? |
| | ● 2-74: | Designer: For example, you can see the corresponding circuit diagram after connecting the physical diagram, or vice versa |
| | ● 2-77: | Designer: Hope to figure out electromagnetism before the end of the year |
| | ● 2-79: | Designer: With bidirectional conversion, you can directly do problems in the application |
| | ● 2-81: | Designer: For example, see the circuit diagram to connect the physical diagram, or vice versa |
| | ● 2-82: | User: Don't aim for completeness, it should be categorized and refined one by one |
| | ● 2-84: | Designer: Physics mainly focuses on electricity and mechanics; other directions don't have much room for free experiments before high school. In the future, we might do some demonstration experiments based on everyone's needs, but it won't be as open as electricity |
| | ● 2-87: | Designer: We'll see if there's a suitable opportunity to do one later |
| | ● 2-88: | Designer: For electricity, we plan to create a function for simulated experiment assessment; it will add many experimental details, such as zero adjustment (of course, only useful in specific modes, otherwise it would be cumbersome to use regularly) |
| | ● 2-90: | Designer: When we are ready to start, we will consult everyone in the group |
| | ● 2-91: | Designer: Consulting the teachers in the group: which type of intersection is used in the circuit diagrams in the current textbooks? [Image] |
| | ● 2-92: | Designer: [Image] This one, right? |
| | ● 2-93: | User: Yes, the common one is still the old style |
| | ● 2-94: | Designer: Uh... left side or right side |
| | ● 2-96: | User: Left side |
| | ● 2-97: | Designer: Okay. |
| | ● 2-99: | User: But it's better to use the right side for non-crossing |
| | ● 2-100: | Designer: So, use both? |
| | ● 2-101: | Designer: This version of the circuit diagram is for testing only... You can drag the Editor because the auto-layout algorithm is not very stable and may produce some odd results |
| | ● 2-103: | Designer: Thank you all for your support. We will do better! |

| past incident | ● 2-41: | User: Generally not. Ever since an adult image popped up during a major city-level open class, the school has disabled the network on classroom computers [Emoji]. |
|---|---|---|
| pc version | ● 2-18: | User: How do you use the PC version? |
| personal interest | ● 2-86: | Designer: I myself have some interest in chemistry |
| personalized assistance | ● 2-69: | Designer: Oh, please message me privately and tell me what you roughly need |
| physics concepts | ● 2-8: | User: First, let's pay homage to the experts, then I'll make a small request. Could you create a dynamic demonstration of mechanical waves and mechanical vibrations, such as the propagation of transverse and longitudinal waves, wave superposition, diffraction, and interference? Also, for optical experiments, it would be great to have optical benches, single slits, double slits, and polarizers to demonstrate optical experiments. |
| platform clarification | ● 2-12: | User: [Emoji] Hello everyone, may I ask where I can download the PC / interactive whiteboard version of Physics Lab? |
| platform limitation | ● 2-20: | Designer: It does not support the XP system. |
| platform specific communication | ● 2-122: | Designer: The one in the group is the PC version |
| platform specific information | ● 2-49: | Designer: Version 1.0.1 of Physics Lab. It is expected to take another one or two days in the Apple market (waiting for review). |
| | ● 2-54: | Designer: You can download Windows in the group files. |
| platform specific update | ● 2-114: | Designer: The Apple Store is still reviewing, let's upload the PC version first |
| | ● 2-118: | Designer: iOS is waiting for review~ |
| | ● 2-123: | Designer: The Android version is expected to update tonight |
| platform specification | ● 2-2: | User: PC version? |
| politeness | ● 2-52: | User: Excuse me. |
| positive feedback | ● 2-46: | User: The simulation effect of this software is really good. When I used it in class yesterday, the students were amazed. |
| | ● 2-76: | User: Oh, this is nice |
| | ● 2-78: | User: The 3D effect of your software is very good |
| | ● 2-80: | User: It's already starting to take shape [Emoji] |
| | ● 2-89: | User: This idea is really good |
| positive interaction | ● 2-24: | User: Thank you. |
| | ● 2-37: | User: @Designer Yes, yes. |
| potential update | ● 2-27: | Designer: XP compatibility might need to be looked into later... probably need to install a virtual machine. |
| practical application | ● 2-26: | User: This software is great! It's quite practical, unlike some software that tries to be comprehensive but ends up being inconvenient to use. |
| private messaging | ● 2-69: | Designer: Oh, please message me privately and tell me what you roughly need |
| problem inquiry | ● 2-65: | Designer: What problems did you encounter during use? |
| problem solving | ● 2-25: | Designer: No need to be polite, if you encounter any problems during use, you can directly mention them in the group. |
| product improvement | ● 2-15: | Designer: From the next update, dry batteries will no longer be ideal components (you can remove the internal resistance to simulate). |
| product iteration | ● 2-29: | Designer: Hello everyone~ Version 1.01 has been released in the group files, the updates are as follows: |
| product update | ● 2-16: | Designer: The student power supply supports both DC and AC and is an ideal component. |
| professional engagement | ● 2-4: | User: Yes. |
| progress update | ● 2-95: | Designer: We are working on this part |
| project management | ● 2-9: | Designer: Hmm... We will probably prioritize completing the electrical section first, then magnetism, and then other parts. |
| project planning | ● 2-10: | Designer: Before starting mechanics, we will gather opinions again~ Otherwise, I'm afraid I won't remember everything. |
| quick response | ● 2-3: | Designer: I'll upload one now... Are you a teacher? |
| real time communication | ● 2-3: | Designer: I'll upload one now... Are you a teacher? |
| reassurance | ● 2-109: | Designer: Don't worry, don't worry, it will come, just optimizing the circuit diagram one last time. |
| release process | ● 2-114:<br>● 2-118: | Designer: The Apple Store is still reviewing, let's upload the PC version first<br>Designer: iOS is waiting for review~ |
| release schedule | ● 2-105: | Designer: There will be an update. Submitting to the app store / releasing the Android version next Monday, and releasing the Windows version over the weekend. |
| research and development | ● 2-71: | Designer: Actually, we are researching the bidirectional conversion between physical diagrams and circuit diagrams |

| resource sharing | ● 2-6: | Designer: Hello everyone~ The development plan and PC download address are in the group announcement. |
|---|---|---|
| resource utilization | ● 2-13: | User: I saw the group files, thank you. |
| school implementation | ● 2-67: | User: Mainly, the school is building an information-based school |
| security concern | ● 2-41: | User: Generally not. Ever since an adult image popped up during a major city-level open class, the school has disabled the network on classroom computers [Emoji]. |
| self sufficiency | ● 2-13: | User: I saw the group files, thank you. |
| simulated experiment | ● 2-88: | Designer: For electricity, we plan to create a function for simulated experiment assessment; it will add many experimental details, such as zero adjustment (of course, only useful in specific modes, otherwise it would be cumbersome to use regularly) |
| simulation accuracy | ● 2-15: | Designer: From the next update, dry batteries will no longer be ideal components (you can remove the internal resistance to simulate). |
| simulation enhancement | ● 2-16: | Designer: The student power supply supports both DC and AC and is an ideal component. |
| soft keyboard consideration | ● 2-33: | Designer: Speaking of which, doesn't that mean every place where numbers are input should have a soft keyboard? |
| software access | ● 2-51: | User: How to download? |
| software effectiveness | ● 2-46: | User: The simulation effect of this software is really good. When I used it in class yesterday, the students were amazed. |
| software release | ● 2-49: | Designer: Version 1.0.1 of Physics Lab. It is expected to take another one or two days in the Apple market (waiting for review). |
| software usability | ● 2-26: | User: This software is great! It's quite practical, unlike some software that tries to be comprehensive but ends up being inconvenient to use. |
| student engagement | ● 2-46: | User: The simulation effect of this software is really good. When I used it in class yesterday, the students were amazed. |
| subject expansion | ● 2-73:<br>● 2-87: | User: Can you also include mechanics experiments?<br>Designer: We'll see if there's a suitable opportunity to do one later |

| subject focus | • 2-84: | Designer: Physics mainly focuses on electricity and mechanics; other directions don't have much room for free experiments before high school. In the future, we might do some demonstration experiments based on everyone's needs, but it won't be as open as electricity |
|---|---|---|
| subject specific tools | • 2-63: | Designer: There will be: multimeter; powered solenoid; semiconductor capacitor; support for conversion to ideal ammeter (more convenient for problem-solving and middle school teaching) |
| | • 2-75: | Designer: Mechanics will have to wait until electromagnetism is figured out; it will take some more time |
| system compatibility | • 2-20: | Designer: It does not support the XP system. |
| | • 2-22: | User: Win7 should be fine. |
| | • 2-27: | Designer: XP compatibility might need to be looked into later... probably need to install a virtual machine. |
| system compatibility confirmation | • 2-23: | Designer: 7 is okay. |
| system limitations | • 2-21: | User: Dizzy, the computer system for teachers at our school is quite old. |
| system requirements | • 2-28: | Designer: Theoretically, it should be compatible (but the machine itself shouldn't be too old, probably from 2008 onwards). |
| teacher identity | • 2-47: | Designer: Haha. Are you a middle school or high school teacher? |
| teacher role | • 2-4: | User: Yes. |
| teaching experience | • 2-48: | User: I taught high school for seven years, and now I've been teaching middle school for eight years. |
| technical clarification | • 2-34: | Designer: Touch screens have their own on-screen keyboards. |
| technical consideration | • 2-27: | Designer: XP compatibility might need to be looked into later... probably need to install a virtual machine. |
| | • 2-40: | Designer: Does the class have internet? |
| technical constraints | • 2-44: | User: Most schools will disable the network on classroom computers to prevent students from going online or to avoid various software auto-downloads from slowing down the computers. |
| technical detail | • 2-15: | Designer: From the next update, dry batteries will no longer be ideal components (you can remove the internal resistance to simulate). |
| | • 2-16: | Designer: The student power supply supports both DC and |

| technical detail | | AC and is an ideal component. |
|---|---|---|
| | ● 2-28: | Designer: Theoretically, it should be compatible (but the machine itself shouldn't be too old, probably from 2008 onwards). |
| | ● 2-35: | Designer: But you can't bring it up without touching the input, that's probably the case. |
| technical details | ● 2-30: | Designer: New Features |
| | | Electronic components will be damaged after a short process, rather than immediately. |
| | | Clearing the desktop will now display a confirmation interface. |
| | | Supports undoing the creation and deletion of wires and components. |
| | | Appliances now display the effective value of alternating current. |
| | | (PC) You can now exit the application using the Esc key. |
| | | New Components |
| | | Added a sensitive ammeter. |
| | | Added a student power supply (ideal AC/DC power supply). |
| | | Adjustments to Component Properties |
| | | Batteries now have adjustable internal resistance and are no longer ideal power sources. |
| | | Incandescent bulbs now have volt-ampere characteristic parameters and are no longer ideal resistors. |
| | | The resistance law experimenter now uses real formulas for calculations, with adjustable parameters. |
| | | Hidden terminal blocks 3 and 4. |
| | | Bug Fixes |
| | | Fixed an issue with unit conversion in Editor properties. |
| | | Terminal arrows no longer show jumping animations. |
| | | There may have been calculation errors with certain circuit connections. |
| technical support | ● 2-1: | Designer: @Morning Tea Moonlight How can I upload a high-definition, uncensored version of this crappy avatar? |
| | ● 2-20: | Designer: It does not support the XP system. |
| | ● 2-23: | Designer: 7 is okay. |
| technical support request | ● 2-18: | User: How do you use the PC version? |
| technical troubleshooting | ● 2-22: | User: Win7 should be fine. |
| testing phase | • 2-101: | Designer: This version of the circuit diagram is for testing only... You can drag the Editor because the auto-layout algorithm is not very stable and may produce some odd results |
| time sensitivity | • 2-50: | Designer: Sorry for the late-night disturbance :) |
| touch screen consideration | • 2-31: | User: It is recommended to add an exit button function to the PC version. Many regions now use all-in-one touch screen machines without physical keyboards. |
| touch screen functionality | • 2-34:<br>• 2-35: | Designer: Touch screens have their own on-screen keyboards.<br>Designer: But you can't bring it up without touching the input, that's probably the case. |
| troubleshooting | • 2-65: | Designer: What problems did you encounter during use? |
| update announcement | • 2-105: | Designer: There will be an update. Submitting to the app store / releasing the Android version next Monday, and releasing the Windows version over the weekend. |
| update clarification | • 2-122: | Designer: The one in the group is the PC version |
| update confirmation | • 2-126: | User: Updated |
| update description | • 2-106: | Designer: The update is quite large... |
| update inquiry | • 2-60:<br>• 2-104: | User: Are there any other updates recently?<br>User: Will there be an update this week? [Emoji] |
| update notification | • 2-117: | Designer: The new version has already been sent in the group |
| update planning | • 2-32: | Designer: Hmm... I'll add it in the next update. PC updates are usually the fastest, so there will probably be another round this weekend. |
| update schedule | • 2-123: | Designer: The Android version is expected to update tonight |
| update status inquiry | • 2-125: | User: Has the Android version not been updated yet? |
| usability | • 2-66: | Designer: We try to design it so that it can be used without additional instructions, or we will provide some prompts when you open it for the first time based on feedback |
| usability discussion | • 2-33: | Designer: Speaking of which, doesn't that mean every place where numbers are input should have a soft keyboard? |
| usability improvement | ● 2-31: | User: It is recommended to add an exit button function to the PC version. Many regions now use all-in-one touch screen machines without physical keyboards. |
| | ● 2-38: | User: If there could be an export function, or the ability to save or import experiments, it would be convenient. We could set up the parameters in the office and directly import them in class. |
| usability workaround | ● 2-36: | Designer: Last time I tried it in middle school, there's a trick where you click the link button in the top right, and when the browser pops up, it's no longer full screen... = = |
| usage inquiry | ● 2-18: | User: How do you use the PC version? |
| usage instruction | ● 2-124: | Designer: After connecting the student power supply, you need to turn on the switch |
| user acknowledgment | ● 2-120: | User: ok |
| user adjustment | ● 2-22: | User: Win7 should be fine. |
| user anticipation | ● 2-104: | User: Will there be an update this week? [Emoji] |
| | ● 2-125: | User: Has the Android version not been updated yet? |
| user appreciation | ● 2-11: | User: Okay, okay~ Thank you for your hard work. |
| | ● 2-112: | User: Yes, the spirit of craftsmanship that strives for perfection 🤔 |
| user assistance request | ● 2-51: | User: How to download? |
| | ● 2-52: | User: Excuse me. |
| user background | ● 2-48: | User: I taught high school for seven years, and now I've been teaching middle school for eight years. |
| user background inquiry | ● 2-47: | Designer: Haha. Are you a middle school or high school teacher? |
| user communication | ● 2-95: | Designer: We are working on this part |
| | ● 2-97: | Designer: Okay. |
| | ● 2-105: | Designer: There will be an update. Submitting to the app store / releasing the Android version next Monday, and releasing the Windows version over the weekend. |
| | ● 2-106: | Designer: The update is quite large... |
| | ● 2-109: | Designer: Don't worry, don't worry, it will come, just optimizing the circuit diagram one last time. |
| | ● 2-114: | Designer: The Apple Store is still reviewing, let's upload the PC version first |
| | ● 2-118: | Designer: iOS is waiting for review~ |
| | ● 2-123: | Designer: The Android version is expected to update tonight |
| | ● 2-126: | User: Updated |

| | |
|---|---|
| user confirmation | • 2-37: User: @Designer Yes, yes.<br>• 2-93: User: Yes, the common one is still the old style<br>• 2-96: User: Left side |
| user engagement | • 2-51: User: How to download?<br>• 2-55: Designer: We will also update the multimeter and powered solenoid this week. If there are any other features, characteristics, or electronic components you hope to include in the production plan, please leave a message in the group, and we will consider it as much as possible. The previously mentioned feature of saving experiments to the cloud/local is also in the plan. Thank you, everyone!<br>• 2-58: User: Hello everyone<br>• 2-60: User: Are there any other updates recently?<br>• 2-76: User: Oh, this is nice<br>• 2-78: User: The 3D effect of your software is very good<br>• 2-80: User: It's already starting to take shape [Emoji]<br>• 2-89: User: This idea is really good<br>• 2-103: Designer: Thank you all for your support. We will do better!<br>• 2-117: Designer: The new version has already been sent in the group |
| user etiquette | • 2-52: User: Excuse me. |
| user expectation management | • 2-62: Designer: This is quite complex, so it will take more time... Hopefully, it can be released this week<br>• 2-77: Designer: Hope to figure out electromagnetism before the end of the year |
| user experience | • 2-36: Designer: Last time I tried it in middle school, there's a trick where you click the link button in the top right, and when the browser pops up, it's no longer full screen... = =<br>• 2-111: Designer: It's still better to make the auto-generated one as good as possible |
| user expression | • 2-57: User: [Emoji][Emoji] |
| user feedback | • 2-66: Designer: We try to design it so that it can be used without additional instructions, or we will provide some prompts when you open it for the first time based on feedback<br>• 2-70: User: This is good, I hope it can be very convenient to draw circuit diagrams, and I also hope there is a function to hide the background grid with one click, which is convenient for us teachers to take screenshots for test papers. Currently, there is no software that makes it easy to draw circuit diagrams for exam questions<br>• 2-71: Designer: Actually, we are researching the bidirectional conversion between physical diagrams and circuit diagrams<br>• 2-72: Designer: Of course, manual editing will also be allowed, but |

| | | |
|---|---|---|
| user feedback | | it might be a bit later |
| | ● 2-73: | User: Can you also include mechanics experiments? |
| | ● 2-75: | Designer: Mechanics will have to wait until electromagnetism is figured out; it will take some more time |
| | ● 2-82: | User: Don't aim for completeness, it should be categorized and refined one by one |
| | ● 2-84: | Designer: Physics mainly focuses on electricity and mechanics; other directions don't have much room for free experiments before high school. In the future, we might do some demonstration experiments based on everyone's needs, but it won't be as open as electricity |
| | ● 2-85: | User: Indeed |
| | ● 2-88: | Designer: For electricity, we plan to create a function for simulated experiment assessment; it will add many experimental details, such as zero adjustment (of course, only useful in specific modes, otherwise it would be cumbersome to use regularly) |
| | ● 2-90: | Designer: When we are ready to start, we will consult everyone in the group |
| | ● 2-91: | Designer: Consulting the teachers in the group: which type of intersection is used in the circuit diagrams in the current textbooks? [Image] |
| | ● 2-92: | Designer: [Image] This one, right? |
| | ● 2-93: | User: Yes, the common one is still the old style |
| | ● 2-98: | Designer: Make it simpler... |
| | ● 2-100: | Designer: So, use both? |
| user feedback request | ● 2-17: | Designer: Okay. Please give more suggestions! |
| user feedback response | ● 2-9: | Designer: Hmm... We will probably prioritize completing the electrical section first, then magnetism, and then other parts. |
| | ● 2-32: | Designer: Hmm... I'll add it in the next update. PC updates are usually the fastest, so there will probably be another round this weekend. |
| user feedback solicitation | ● 2-55: | Designer: We will also update the multimeter and powered solenoid this week. If there are any other features, characteristics, or electronic components you hope to include in the production plan, please leave a message in the group, and we will consider it as much as possible. The previously mentioned feature of saving experiments to the cloud/local is also in the plan. Thank you, everyone! |
| user frustration | ● 2-21: | User: Dizzy, the computer system for teachers at our school is quite old. |
| user gratitude | ● 2-24: | User: Thank you. |

| | |
|---|---|
| user guidance | • 2-15:  Designer: From the next update, dry batteries will no longer be ideal components (you can remove the internal resistance to simulate).<br><br>• 2-20:  Designer: It does not support the XP system.<br><br>• 2-28:  Designer: Theoretically, it should be compatible (but the machine itself shouldn't be too old, probably from 2008 onwards).<br><br>• 2-30:  Designer: New Features<br>Electronic components will be damaged after a short process, rather than immediately.<br>Clearing the desktop will now display a confirmation interface.<br>Supports undoing the creation and deletion of wires and components.<br>Appliances now display the effective value of alternating current.<br>(PC) You can now exit the application using the Esc key.<br><br>New Components<br>Added a sensitive ammeter.<br>Added a student power supply (ideal AC/DC power supply).<br><br>Adjustments to Component Properties<br>Batteries now have adjustable internal resistance and are no longer ideal power sources.<br>Incandescent bulbs now have volt-ampere characteristic parameters and are no longer ideal resistors.<br>The resistance law experimenter now uses real formulas for calculations, with adjustable parameters.<br>Hidden terminal blocks 3 and 4.<br><br>Bug Fixes<br>Fixed an issue with unit conversion in Editor properties.<br>Terminal arrows no longer show jumping animations.<br>There may have been calculation errors with certain circuit connections.<br><br>• 2-34:  Designer: Touch screens have their own on-screen keyboards.<br><br>• 2-54:  Designer: You can download Windows in the group files.<br><br>• 2-63:  Designer: There will be: multimeter; powered solenoid; semiconductor capacitor; support for conversion to ideal ammeter (more convenient for problem-solving and middle school teaching)<br><br>• 2-74:  Designer: For example, you can see the corresponding circuit diagram after connecting the physical diagram, or vice versa<br><br>• 2-79:  Designer: With bidirectional conversion, you can directly do problems in the application |

| | | |
|---|---|---|
| user guidance | ● 2-81: | Designer: For example, see the circuit diagram to connect the physical diagram, or vice versa |
| | ● 2-101: | Designer: This version of the circuit diagram is for testing only... You can drag the Editor because the auto-layout algorithm is not very stable and may produce some odd results |
| | ● 2-110: | Designer: Although you can manually adjust the layout |
| | ● 2-122: | Designer: The one in the group is the PC version |
| user identity | ● 2-4: | User: Yes. |
| user input | ● 2-10: | Designer: Before starting mechanics, we will gather opinions again~ Otherwise, I'm afraid I won't remember everything. |
| user inquiry | ● 2-2: | User: PC version? |
| user interaction | ● 2-35: | Designer: But you can't bring it up without touching the input, that's probably the case. |
| | ● 2-94: | Designer: Uh... left side or right side |
| user interface | ● 2-1: | Designer: @Morning Tea Moonlight How can I upload a high-definition, uncensored version of this crappy avatar? |
| user manual inquiry | ● 2-64: | User: Is there a user manual? |
| user manual request | ● 2-68: | User: Need the user manual for this software |
| user needs | ● 2-67: | User: Mainly, the school is building an information-based school |
| | ● 2-68: | User: Need the user manual for this software |
| user reassurance | ● 2-23: | Designer: 7 is okay. |
| user response | ● 2-57: | User: [Emoji][Emoji] |
| user role inquiry | ● 2-3: | Designer: I'll upload one now... Are you a teacher? |
| user satisfaction | ● 2-26: | User: This software is great! It's quite practical, unlike some software that tries to be comprehensive but ends up being inconvenient to use. |
| user suggestion | ● 2-31: | User: It is recommended to add an exit button function to the PC version. Many regions now use all-in-one touch screen machines without physical keyboards. |
| | ● 2-99: | User: But it's better to use the right side for non-crossing |
| user suggestions | ● 2-7: | Designer: If you have any suggestions or requirements, feel free to bring them up. |
| user support | ● 2-25: | Designer: No need to be polite, if you encounter any problems during use, you can directly mention them in the group. |
| | ● 2-64: | User: Is there a user manual? |

| user support | • 2-65:   Designer: What problems did you encounter during use? |
|---|---|
| | • 2-69:   Designer: Oh, please message me privately and tell me what you roughly need |
| | • 2-124:  Designer: After connecting the student power supply, you need to turn on the switch |
| user tips | • 2-115:  Designer: v1.0.2 - 17/11/29 |
| | # New Features |
| | Multi-language support: Japanese, German, French. |
| | Preliminary support for converting circuit components to circuit diagrams. |
| | Support for augmented reality mode on the latest iOS devices. |
| | More realistic 3D appearance for various meters. |
| | |
| | # Update Preview |
| | The next version will provide multimeter and meter zeroing. |
| | The next version will provide AC waveform display functionality. |
| | The next version will provide capacitors and energized solenoids. |
| | |
| | # Component Property Adjustments |
| | Dry cells can now be ideal power sources. |
| | All meters can now switch between ideal/realistic modes. |
| | |
| | # Special Tips |
| | The Esc shortcut key will now minimize the window. |
| | The Ctrl+E shortcut key can switch between exam and normal modes. |
| version clarification | • 2-2:    User: PC version? |
| version details | • 2-115:  Designer: v1.0.2 - 17/11/29 |
| | # New Features |
| | Multi-language support: Japanese, German, French. |
| | Preliminary support for converting circuit components to circuit diagrams. |
| | Support for augmented reality mode on the latest iOS devices. |
| | More realistic 3D appearance for various meters. |
| | |
| | # Update Preview |
| | The next version will provide multimeter and meter zeroing. |
| | The next version will provide AC waveform display functionality. |
| version details | The next version will provide capacitors and energized solenoids.<br><br># Component Property Adjustments<br>Dry cells can now be ideal power sources.<br>All meters can now switch between ideal/realistic modes.<br><br># Special Tips<br>The Esc shortcut key will now minimize the window.<br>The Ctrl+E shortcut key can switch between exam and normal modes. |
| version update announcement | • 2-29:   Designer: Hello everyone~ Version 1.01 has been released in the group files, the updates are as follows:<br>• 2-49:   Designer: Version 1.0.1 of Physics Lab. It is expected to take another one or two days in the Apple market (waiting for review). |
| visual confirmation | • 2-92:   Designer: [Image] This one, right? |
| visual design appreciation | • 2-78:   User: The 3D effect of your software is very good |
| visual update | • 2-61:   Designer: [Image] Updates in preparation |
| workaround sharing | • 2-36:   Designer: Last time I tried it in middle school, there's a trick where you click the link button in the top right, and when the browser pops up, it's no longer full screen... = = |

## 4.4 Item-level Coding w/ Verb Phrases

| Label | Examples |
|---|---|
| acknowledge assistance | • 2-24:   User: Thank you. |
| acknowledge craftsmanship | • 2-112:   User: Yes, the spirit of craftsmanship that strives for perfection 🤔 |
| acknowledge designer effort | • 2-11:   User: Okay, okay~ Thank you for your hard work. |
| acknowledge feature | • 2-76:   User: Oh, this is nice |
| acknowledge feedback | • 2-83:   Designer: Hmm.<br>• 2-97:   Designer: Okay. |
| acknowledge finding resources | • 2-13:   User: I saw the group files, thank you. |
| acknowledge information | • 2-120:   User: ok |
| acknowledge limitations | • 2-27:   Designer: XP compatibility might need to be looked into later... probably need to install a virtual machine. |

| | | |
|---|---|---|
| acknowledge need for local export | • 2-43: | Designer: It seems that local export needs to be supported. |
| acknowledge potential issue | • 2-35: | Designer: But you can't bring it up without touching the input, that's probably the case. |
| acknowledge suggestion | • 2-32: | Designer: Hmm... I'll add it in the next update. PC updates are usually the fastest, so there will probably be another round this weekend. |
| acknowledge user input | • 2-45: | Designer: Hmm hmm. |
| adapt design plan | • 2-43: | Designer: It seems that local export needs to be supported. |
| adapt to platform specific constraints | • 2-114: | Designer: The Apple Store is still reviewing, let's upload the PC version first |
| address feedback process | • 2-66: | Designer: We try to design it so that it can be used without additional instructions, or we will provide some prompts when you open it for the first time based on feedback |
| address security and performance concerns | • 2-44: | User: Most schools will disable the network on classroom computers to prevent students from going online or to avoid various software auto-downloads from slowing down the computers. |
| address technical limitation | • 2-20: | Designer: It does not support the XP system. |
| address usability concern | • 2-34: | Designer: Touch screens have their own on-screen keyboards. |
| advise on development strategy | • 2-82: | User: Don't aim for completeness, it should be categorized and refined one by one |
| affirm planned feature | • 2-39: | Designer: There will be. |
| agree with focus areas | • 2-85: | User: Indeed |
| agree with suggestion | • 2-37: | User: @Designer Yes, yes. |
| announce component change | • 2-15: | Designer: From the next update, dry batteries will no longer be ideal components (you can remove the internal resistance to simulate). |
| announce future consultation | • 2-90: | Designer: When we are ready to start, we will consult everyone in the group |
| announce new version release | • 2-29: | Designer: Hello everyone~ Version 1.01 has been released in the group files, the updates are as follows: |
| announce resource availability | • 2-6: | Designer: Hello everyone~ The development plan and PC download address are in the group announcement. |
| announce software version update | ● 2-49: | Designer: Version 1.0.1 of Physics Lab. It is expected to take another one or two days in the Apple market (waiting for review). |
| announce upcoming features | ● 2-55: | Designer: We will also update the multimeter and powered solenoid this week. If there are any other features, characteristics, or electronic components you hope to include in the production plan, please leave a message in the group, and we will consider it as much as possible. The previously mentioned feature of saving experiments to the cloud/local is also in the plan. Thank you, everyone! |
| apologize for interruption | ● 2-52: | User: Excuse me. |
| apologize for late message | ● 2-50: | Designer: Sorry for the late-night disturbance :) |
| appreciate design effort | ● 2-112: | User: Yes, the spirit of craftsmanship that strives for perfection 🤭 |
| ask for download instructions | ● 2-51: | User: How to download? |
| ask for resource location | ● 2-12: | User: [Emoji] Hello everyone, may I ask where I can download the PC / interactive whiteboard version of Physics Lab? |
| ask for technical assistance | ● 2-1: | Designer: @Morning Tea Moonlight How can I upload a high-definition, uncensored version of this crappy avatar? |
| build community rapport | ● 2-103: | Designer: Thank you all for your support. We will do better! |
| clarify existing functionality | ● 2-34: | Designer: Touch screens have their own on-screen keyboards. |
| clarify preference | ● 2-96: | User: Left side |
| clarify version details | ● 2-122: | Designer: The one in the group is the PC version |
| close inquiry loop | ● 2-13: | User: I saw the group files, thank you. |
| commit to future update | ● 2-32: | Designer: Hmm... I'll add it in the next update. PC updates are usually the fastest, so there will probably be another round this weekend. |
| commit to improvement | ● 2-103: | Designer: Thank you all for your support. We will do better! |
| communicate development process | ● 2-9: | Designer: Hmm... We will probably prioritize completing the electrical section first, then magnetism, and then other parts. |
| communicate development progress | ● 2-105: | Designer: There will be an update. Submitting to the app store / releasing the Android version next Monday, and releasing the Windows version over the weekend. |
| communicate long term benefits | ● 2-113: | Designer: This belongs to the kind of feature that, once done, will ensure long-term stability... Adding various components is actually simpler |
| communicate new features and improvements | ● 2-115: | Designer: v1.0.2 - 17/11/29<br># New Features<br>Multi-language support: Japanese, German, French.<br>Preliminary support for converting circuit components to circuit diagrams.<br>Support for augmented reality mode on the latest iOS devices.<br>More realistic 3D appearance for various meters.<br><br># Update Preview<br>The next version will provide multimeter and meter zeroing.<br>The next version will provide AC waveform display functionality.<br>The next version will provide capacitors and energized solenoids.<br><br># Component Property Adjustments<br>Dry cells can now be ideal power sources.<br>All meters can now switch between ideal/realistic modes.<br><br># Special Tips<br>The Esc shortcut key will now minimize the window.<br>The Ctrl+E shortcut key can switch between exam and normal modes. |
| communicate ongoing optimization | ● 2-109: | Designer: Don't worry, don't worry, it will come, just optimizing the circuit diagram one last time. |
| communicate progress | ● 2-123: | Designer: The Android version is expected to update tonight |
| compare with other software | ● 2-26: | User: This software is great! It's quite practical, unlike some software that tries to be comprehensive but ends up being inconvenient to use. |
| compliment software | ● 2-78: | User: The 3D effect of your software is very good |
| confirm alignment with user needs | ● 2-39: | Designer: There will be. |
| confirm common standard | ● 2-93: | User: Yes, the common one is still the old style |
| confirm compatibility | ● 2-23: | Designer: 7 is okay. |
| confirm future feature | ● 2-72: | Designer: Of course, manual editing will also be allowed, but it might be a bit later |

| confirm professional role | • 2-4: | User: Yes. |
|---|---|---|
| confirm understanding | • 2-37:<br>• 2-92:<br>• 2-97: | User: @Designer Yes, yes.<br>Designer: [Image] This one, right?<br>Designer: Okay. |
| confirm update | • 2-105: | Designer: There will be an update. Submitting to the app store / releasing the Android version next Monday, and releasing the Windows version over the weekend. |
| confirm update completion | • 2-126: | User: Updated |
| consider future compatibility | • 2-27: | Designer: XP compatibility might need to be looked into later... probably need to install a virtual machine. |
| consider implementation context | • 2-40: | Designer: Does the class have internet? |
| consider implications of touch screen use | • 2-33: | Designer: Speaking of which, doesn't that mean every place where numbers are input should have a soft keyboard? |
| consider usability | • 2-98:<br>• 2-99: | Designer: Make it simpler...<br>User: But it's better to use the right side for non-crossing |
| consider user experience | • 2-35: | Designer: But you can't bring it up without touching the input, that's probably the case. |
| consider user feedback | • 2-43: | Designer: It seems that local export needs to be supported. |
| describe component updates | • 2-30: | Designer: New Features<br>Electronic components will be damaged after a short process, rather than immediately.<br>Clearing the desktop will now display a confirmation interface.<br>Supports undoing the creation and deletion of wires and components.<br>Appliances now display the effective value of alternating current.<br>(PC) You can now exit the application using the Esc key.<br><br>New Components<br>Added a sensitive ammeter.<br>Added a student power supply (ideal AC/DC power supply).<br><br>Adjustments to Component Properties<br>Batteries now have adjustable internal resistance and are no longer ideal power sources. |
| describe component updates | | Incandescent bulbs now have volt-ampere characteristic parameters and are no longer ideal resistors.<br>The resistance law experimenter now uses real formulas for calculations, with adjustable parameters.<br>Hidden terminal blocks 3 and 4.<br><br>Bug Fixes<br>Fixed an issue with unit conversion in Editor properties.<br>Terminal arrows no longer show jumping animations.<br>There may have been calculation errors with certain circuit connections. |
| describe educational needs | ● 2-8: | User: First, let's pay homage to the experts, then I'll make a small request. Could you create a dynamic demonstration of mechanical waves and mechanical vibrations, such as the propagation of transverse and longitudinal waves, wave superposition, diffraction, and interference? Also, for optical experiments, it would be great to have optical benches, single slits, double slits, and polarizers to demonstrate optical experiments. |
| describe educational use case | ● 2-70: | User: This is good, I hope it can be very convenient to draw circuit diagrams, and I also hope there is a function to hide the background grid with one click, which is convenient for us teachers to take screenshots for test papers. Currently, there is no software that makes it easy to draw circuit diagrams for exam questions |
| describe feature importance | ● 2-113: | Designer: This belongs to the kind of feature that, once done, will ensure long-term stability... Adding various components is actually simpler |
| describe future plans | ● 2-84: | Designer: Physics mainly focuses on electricity and mechanics; other directions don't have much room for free experiments before high school. In the future, we might do some demonstration experiments based on everyone's needs, but it won't be as open as electricity |
| describe institutional needs | ● 2-67: | User: Mainly, the school is building an information-based school |
| describe outdated school system | ● 2-21: | User: Dizzy, the computer system for teachers at our school is quite old. |
| describe planned feature | ● 2-88: | Designer: For electricity, we plan to create a function for simulated experiment assessment; it will add many experimental details, such as zero adjustment (of course, only useful in specific modes, otherwise it would be cumbersome to use regularly) |

| | |
|---|---|
| describe update scope | ● 2-106:  Designer: The update is quite large... |
| describe usability features | ● 2-66:  Designer: We try to design it so that it can be used without additional instructions, or we will provide some prompts when you open it for the first time based on feedback |
| describe usage scenario | ● 2-79:  Designer: With bidirectional conversion, you can directly do problems in the application |
| detail educational utility | ● 2-88:  Designer: For electricity, we plan to create a function for simulated experiment assessment; it will add many experimental details, such as zero adjustment (of course, only useful in specific modes, otherwise it would be cumbersome to use regularly) |
| detail new features | ● 2-30:  Designer: New Features<br>Electronic components will be damaged after a short process, rather than immediately.<br>Clearing the desktop will now display a confirmation interface.<br>Supports undoing the creation and deletion of wires and components.<br>Appliances now display the effective value of alternating current.<br>(PC) You can now exit the application using the Esc key.<br><br>New Components<br>Added a sensitive ammeter.<br>Added a student power supply (ideal AC/DC power supply).<br><br>Adjustments to Component Properties<br>Batteries now have adjustable internal resistance and are no longer ideal power sources.<br>Incandescent bulbs now have volt-ampere characteristic parameters and are no longer ideal resistors.<br>The resistance law experimenter now uses real formulas for calculations, with adjustable parameters.<br>Hidden terminal blocks 3 and 4.<br><br>Bug Fixes<br>Fixed an issue with unit conversion in Editor properties.<br>Terminal arrows no longer show jumping animations.<br>There may have been calculation errors with certain circuit connections. |

| detailed update notes | ● 2-115: | Designer: v1.0.2 - 17/11/29 |
|---|---|---|
| | | # New Features |
| | | Multi-language support: Japanese, German, French. |
| | | Preliminary support for converting circuit components to circuit diagrams. |
| | | Support for augmented reality mode on the latest iOS devices. |
| | | More realistic 3D appearance for various meters. |
| | | |
| | | # Update Preview |
| | | The next version will provide multimeter and meter zeroing. |
| | | The next version will provide AC waveform display functionality. |
| | | The next version will provide capacitors and energized solenoids. |
| | | |
| | | # Component Property Adjustments |
| | | Dry cells can now be ideal power sources. |
| | | All meters can now switch between ideal/realistic modes. |
| | | |
| | | # Special Tips |
| | | The Esc shortcut key will now minimize the window. |
| | | The Ctrl+E shortcut key can switch between exam and normal modes. |
| direct users to resources | ● 2-54: | Designer: You can download Windows in the group files. |
| emoji | ● 2-56: | User: [Emoji] |
| | ● 2-107: | User: [Emoji] |
| | ● 2-108: | User: [Emoji] |
| | ● 2-116: | User: [Emoji] |
| | ● 2-119: | User: [Emoji] |
| | ● 2-121: | User: [Emoji] |
| emphasize continuous improvement | ● 2-30: | Designer: New Features |
| | | Electronic components will be damaged after a short process, rather than immediately. |
| | | Clearing the desktop will now display a confirmation interface. |
| | | Supports undoing the creation and deletion of wires and components. |
| | | Appliances now display the effective value of alternating current. |
| | | (PC) You can now exit the application using the Esc key. |
| | | |
| | | New Components |
| emphasize continuous improvement | | Added a sensitive ammeter.<br>Added a student power supply (ideal AC/DC power supply).<br><br>Adjustments to Component Properties<br>Batteries now have adjustable internal resistance and are no longer ideal power sources.<br>Incandescent bulbs now have volt-ampere characteristic parameters and are no longer ideal resistors.<br>The resistance law experimenter now uses real formulas for calculations, with adjustable parameters.<br>Hidden terminal blocks 3 and 4.<br><br>Bug Fixes<br>Fixed an issue with unit conversion in Editor properties.<br>Terminal arrows no longer show jumping animations.<br>There may have been calculation errors with certain circuit connections. |
| emphasize convenience | ● 2-38: | User: If there could be an export function, or the ability to save or import experiments, it would be convenient. We could set up the parameters in the office and directly import them in class. |
| emphasize importance | ● 2-68: | User: Need the user manual for this software |
| emphasize participatory design | ● 2-90: | Designer: When we are ready to start, we will consult everyone in the group |
| emphasize quality | ● 2-111: | Designer: It's still better to make the auto-generated one as good as possible |
| encourage community feedback | ● 2-17: | Designer: Okay. Please give more suggestions! |
| encourage community participation | ● 2-55: | Designer: We will also update the multimeter and powered solenoid this week. If there are any other features, characteristics, or electronic components you hope to include in the production plan, please leave a message in the group, and we will consider it as much as possible. The previously mentioned feature of saving experiments to the cloud/local is also in the plan. Thank you, everyone! |
| encourage problem reporting | ● 2-25: | Designer: No need to be polite, if you encounter any problems during use, you can directly mention them in the group. |
| encourage user suggestions | ● 2-7: | Designer: If you have any suggestions or requirements, feel free to bring them up. |
| engage community | ● 2-29: | Designer: Hello everyone~ Version 1.01 has been released in the group files, the updates are as follows: |
| engage community in design process | ● 2-91: | Designer: Consulting the teachers in the group: which type of intersection is used in the circuit diagrams in the current textbooks? [Image] |
| engage community member | ● 2-1: | Designer: @Morning Tea Moonlight How can I upload a high-definition, uncensored version of this crappy avatar? |
| engage in participatory design | ● 2-8: | User: First, let's pay homage to the experts, then I'll make a small request. Could you create a dynamic demonstration of mechanical waves and mechanical vibrations, such as the propagation of transverse and longitudinal waves, wave superposition, diffraction, and interference? Also, for optical experiments, it would be great to have optical benches, single slits, double slits, and polarizers to demonstrate optical experiments. |
| engage in problem solving | ● 2-65: | Designer: What problems did you encounter during use? |
| engage users with update | ● 2-49: | Designer: Version 1.0.1 of Physics Lab. It is expected to take another one or two days in the Apple market (waiting for review). |
| engage with community | ● 2-12: | User: [Emoji] Hello everyone, may I ask where I can download the PC / interactive whiteboard version of Physics Lab? |
| | ● 2-57: | User: [Emoji][Emoji] |
| engage with the community | ● 2-1: | Designer: @Morning Tea Moonlight How can I upload a high-definition, uncensored version of this crappy avatar? |
| ensure proper use of new features | ● 2-124: | Designer: After connecting the student power supply, you need to turn on the switch |
| establish credibility | ● 2-48: | User: I taught high school for seven years, and now I've been teaching middle school for eight years. |
| establish identity within the community | ● 2-4: | User: Yes. |
| establish user identity | ● 2-4: | User: Yes. |
| explain bidirectional conversion feature | ● 2-74: | Designer: For example, you can see the corresponding circuit diagram after connecting the physical diagram, or vice versa |
| explain common practice in schools | ● 2-44: | User: Most schools will disable the network on classroom computers to prevent students from going online or to avoid various software auto-downloads from slowing down the computers. |
| explain complexity | ● 2-62: | Designer: This is quite complex, so it will take more time... Hopefully, it can be released this week |
| explain current limitations | ● 2-101: | Designer: This version of the circuit diagram is for testing only... You can drag the Editor because the auto-layout algorithm is not very stable and may produce some odd results |
| explain design philosophy | ● 2-66: | Designer: We try to design it so that it can be used without additional instructions, or we will provide some prompts when you open it for the first time based on feedback |
| explain development timeline | ● 2-75: | Designer: Mechanics will have to wait until electromagnetism is figured out; it will take some more time |
| explain focus areas | ● 2-84: | Designer: Physics mainly focuses on electricity and mechanics; other directions don't have much room for free experiments before high school. In the future, we might do some demonstration experiments based on everyone's needs, but it won't be as open as electricity |
| explain lack of internet | ● 2-41: | User: Generally not. Ever since an adult image popped up during a major city-level open class, the school has disabled the network on classroom computers [Emoji]. |
| explain new component capabilities | ● 2-16: | Designer: The student power supply supports both DC and AC and is an ideal component. |
| express amusement | ● 2-42: | Designer: 😂 |
| express approval | ● 2-76:<br>● 2-80:<br>● 2-89: | User: Oh, this is nice<br>User: It's already starting to take shape [Emoji]<br>User: This idea is really good |
| express approval or acknowledgment | ● 2-57: | User: [Emoji][Emoji] |
| express contemplation | ● 2-83: | Designer: Hmm. |
| express frustration | ● 2-21: | User: Dizzy, the computer system for teachers at our school is quite old. |
| express gratitude | ● 2-11:<br>● 2-13:<br>● 2-24:<br>● 2-103: | User: Okay, okay~ Thank you for your hard work.<br>User: I saw the group files, thank you.<br>User: Thank you.<br>Designer: Thank you all for your support. We will do better! |
| express interest or need | ● 2-2: | User: PC version? |
| express need for specific features | • 2-70: | User: This is good, I hope it can be very convenient to draw circuit diagrams, and I also hope there is a function to hide the background grid with one click, which is convenient for us teachers to take screenshots for test papers. Currently, there is no software that makes it easy to draw circuit diagrams for exam questions |
| express openness | • 2-87: | Designer: We'll see if there's a suitable opportunity to do one later |
| express personal interest | • 2-86: | Designer: I myself have some interest in chemistry |
| express satisfaction | • 2-26: | User: This software is great! It's quite practical, unlike some software that tries to be comprehensive but ends up being inconvenient to use. |
| express understanding | • 2-45:<br>• 2-120: | Designer: Hmm hmm.<br>User: ok |
| facilitate access | • 2-6:<br>• 2-53: | Designer: Hello everyone~ The development plan and PC download address are in the group announcement.<br>Designer: Group sharing. |
| focus on user experience | • 2-31: | User: It is recommended to add an exit button function to the PC version. Many regions now use all-in-one touch screen machines without physical keyboards. |
| foster open dialogue | • 2-25: | Designer: No need to be polite, if you encounter any problems during use, you can directly mention them in the group. |
| gather additional information | • 2-40: | Designer: Does the class have internet? |
| greet community | • 2-58: | User: Hello everyone |
| highlight educational tools | • 2-63: | Designer: There will be: multimeter; powered solenoid; semiconductor capacitor; support for conversion to ideal ammeter (more convenient for problem-solving and middle school teaching) |
| highlight practical application | • 2-79: | Designer: With bidirectional conversion, you can directly do problems in the application |
| highlight practicality | • 2-26: | User: This software is great! It's quite practical, unlike some software that tries to be comprehensive but ends up being inconvenient to use. |
| highlight security concerns | • 2-41: | User: Generally not. Ever since an adult image popped up during a major city-level open class, the school has disabled the network on classroom computers [Emoji]. |
| highlight student engagement | ● 2-46: | User: The simulation effect of this software is really good. When I used it in class yesterday, the students were amazed. |
| highlight update frequency | ● 2-32: | Designer: Hmm... I'll add it in the next update. PC updates are usually the fastest, so there will probably be another round this weekend. |
| hint at future possibilities | ● 2-86: | Designer: I myself have some interest in chemistry |
| identify potential need | ● 2-33: | Designer: Speaking of which, doesn't that mean every place where numbers are input should have a soft keyboard? |
| identify usability issue | ● 2-31: | User: It is recommended to add an exit button function to the PC version. Many regions now use all-in-one touch screen machines without physical keyboards. |
| identify user needs | ● 2-3: | Designer: I'll upload one now... Are you a teacher? |
| image sharing | ● 2-0:<br>● 2-14:<br>● 2-19:<br>● 2-102: | Designer: [Image]<br>Designer: [Image]<br>User: [Image]<br>Designer: [Image] |
| improve avatar quality | ● 2-1: | Designer: @Morning Tea Moonlight How can I upload a high-definition, uncensored version of this crappy avatar? |
| indicate interest or need | ● 2-2: | User: PC version? |
| indicate ongoing work | ● 2-95: | Designer: We are working on this part |
| inform about development plan | ● 2-6: | Designer: Hello everyone~ The development plan and PC download address are in the group announcement. |
| inform about ideal components | ● 2-16: | Designer: The student power supply supports both DC and AC and is an ideal component. |
| inform about ios review status | ● 2-118: | Designer: iOS is waiting for review~ |
| inform about new version release | ● 2-117: | Designer: The new version has already been sent in the group |
| inform about release timeline | ● 2-49: | Designer: Version 1.0.1 of Physics Lab. It is expected to take another one or two days in the Apple market (waiting for review). |
| inform about simulation capability | ● 2-15: | Designer: From the next update, dry batteries will no longer be ideal components (you can remove the internal resistance to simulate). |
| inform about system compatibility | ● 2-20: | Designer: It does not support the XP system. |

| inform about updates | ● 2-29: | Designer: Hello everyone~ Version 1.01 has been released in the group files, the updates are as follows: |
|---|---|---|
| initiate conversation | ● 2-52: | User: Excuse me. |
| initiate interaction | ● 2-58: | User: Hello everyone |
| inquire about classroom internet | ● 2-40: | Designer: Does the class have internet? |
| inquire about specific issues | ● 2-65: | Designer: What problems did you encounter during use? |
| inquire about specific platform | ● 2-2: | User: PC version? |
| inquire about update status | ● 2-125: | User: Has the Android version not been updated yet? |
| inquire about update timeline | ● 2-104: | User: Will there be an update this week? [Emoji] |
| inquire about updates | ● 2-60: | User: Are there any other updates recently? |
| inquire about usage | ● 2-18: | User: How do you use the PC version? |
| inquire about user manual | ● 2-64: | User: Is there a user manual? |
| inquire about user's role | ● 2-3: | Designer: I'll upload one now... Are you a teacher? |
| inquire about user's teaching background | ● 2-47: | Designer: Haha. Are you a middle school or high school teacher? |
| inquire about user's role | ● 2-3: | Designer: I'll upload one now... Are you a teacher? |
| invite feedback | ● 2-7: | Designer: If you have any suggestions or requirements, feel free to bring them up. |
| justify request | ● 2-67: | User: Mainly, the school is building an information-based school |
| list bug fixes | ● 2-30: | Designer: New Features<br>Electronic components will be damaged after a short process, rather than immediately.<br>Clearing the desktop will now display a confirmation interface.<br>Supports undoing the creation and deletion of wires and components.<br>Appliances now display the effective value of alternating current.<br>(PC) You can now exit the application using the Esc key.<br><br>New Components<br>Added a sensitive ammeter.<br>Added a student power supply (ideal AC/DC power supply).<br><br>Adjustments to Component Properties |
| list bug fixes | | Batteries now have adjustable internal resistance and are no longer ideal power sources.<br>Incandescent bulbs now have volt-ampere characteristic parameters and are no longer ideal resistors.<br>The resistance law experimenter now uses real formulas for calculations, with adjustable parameters.<br>Hidden terminal blocks 3 and 4.<br><br>Bug Fixes<br>Fixed an issue with unit conversion in Editor properties.<br>Terminal arrows no longer show jumping animations.<br>There may have been calculation errors with certain circuit connections. |
| list upcoming features | • 2-63: | Designer: There will be: multimeter; powered solenoid; semiconductor capacitor; support for conversion to ideal ammeter (more convenient for problem-solving and middle school teaching) |
| maintain polite communication | • 2-50: | Designer: Sorry for the late-night disturbance :) |
| maintain social norms | • 2-52:<br>• 2-59: | User: Excuse me.<br>Designer: Hello :) |
| maintain user engagement | • 2-10: | Designer: Before starting mechanics, we will gather opinions again~ Otherwise, I'm afraid I won't remember everything. |
| manage expectations | • 2-49:<br><br>• 2-62:<br><br>• 2-72:<br><br>• 2-75:<br><br>• 2-77:<br><br>• 2-84:<br><br><br><br><br>• 2-106: | Designer: Version 1.0.1 of Physics Lab. It is expected to take another one or two days in the Apple market (waiting for review).<br>Designer: This is quite complex, so it will take more time... Hopefully, it can be released this week<br>Designer: Of course, manual editing will also be allowed, but it might be a bit later<br>Designer: Mechanics will have to wait until electromagnetism is figured out; it will take some more time<br>Designer: Hope to figure out electromagnetism before the end of the year<br>Designer: Physics mainly focuses on electricity and mechanics; other directions don't have much room for free experiments before high school. In the future, we might do some demonstration experiments based on everyone's needs, but it won't be as open as electricity<br>Designer: The update is quite large... |

| manage expectations | • 2-115: | Designer: v1.0.2 - 17/11/29<br># New Features<br>Multi-language support: Japanese, German, French.<br>Preliminary support for converting circuit components to circuit diagrams.<br>Support for augmented reality mode on the latest iOS devices.<br>More realistic 3D appearance for various meters.<br><br># Update Preview<br>The next version will provide multimeter and meter zeroing.<br>The next version will provide AC waveform display functionality.<br>The next version will provide capacitors and energized solenoids.<br><br># Component Property Adjustments<br>Dry cells can now be ideal power sources.<br>All meters can now switch between ideal/realistic modes.<br><br># Special Tips<br>The Esc shortcut key will now minimize the window.<br>The Ctrl+E shortcut key can switch between exam and normal modes. |
| --- | --- | --- |
| manage memory and task organization | • 2-10: | Designer: Before starting mechanics, we will gather opinions again~ Otherwise, I'm afraid I won't remember everything. |
| manage platform specific expectations | • 2-118: | Designer: iOS is waiting for review~ |
| manage user expectations | • 2-9:<br><br>• 2-101:<br><br><br><br>• 2-109:<br><br>• 2-122: | Designer: Hmm... We will probably prioritize completing the electrical section first, then magnetism, and then other parts.<br>Designer: This version of the circuit diagram is for testing only... You can drag the Editor because the auto-layout algorithm is not very stable and may produce some odd results<br>Designer: Don't worry, don't worry, it will come, just optimizing the circuit diagram one last time.<br>Designer: The one in the group is the PC version |
| offer personalized support | • 2-69: | Designer: Oh, please message me privately and tell me what you roughly need |
| offer support | • 2-54:<br>• 2-65: | Designer: You can download Windows in the group files.<br>Designer: What problems did you encounter during use? |
| offer to provide resources | ● 2-3: | Designer: I'll upload one now... Are you a teacher? |
| offer to upload resource | ● 2-3: | Designer: I'll upload one now... Are you a teacher? |
| outline development priorities | ● 2-9: | Designer: Hmm... We will probably prioritize completing the electrical section first, then magnetism, and then other parts. |
| plan to gather feedback | ● 2-10: | Designer: Before starting mechanics, we will gather opinions again~ Otherwise, I'm afraid I won't remember everything. |
| praise simulation effect | ● 2-46: | User: The simulation effect of this software is really good. When I used it in class yesterday, the students were amazed. |
| praise software | ● 2-26: | User: This software is great! It's quite practical, unlike some software that tries to be comprehensive but ends up being inconvenient to use. |
| prepare users for significant changes | ● 2-106: | Designer: The update is quite large... |
| prioritize user experience | ● 2-111: | Designer: It's still better to make the auto-generated one as good as possible |
| promote group communication | ● 2-25: | Designer: No need to be polite, if you encounter any problems during use, you can directly mention them in the group. |
| promote participatory design | ● 2-7: | Designer: If you have any suggestions or requirements, feel free to bring them up. |
| propose compromise | ● 2-100: | Designer: So, use both? |
| propose solution | ● 2-22: | User: Win7 should be fine. |
| propose workflow improvement | ● 2-38: | User: If there could be an export function, or the ability to save or import experiments, it would be convenient. We could set up the parameters in the office and directly import them in class. |
| provide alternative suggestion | ● 2-99: | User: But it's better to use the right side for non-crossing |
| provide background information | ● 2-48: | User: I taught high school for seven years, and now I've been teaching middle school for eight years. |
| provide context | ● 2-67: | User: Mainly, the school is building an information-based school |
| provide download instructions | ● 2-54: | Designer: You can download Windows in the group files. |

| provide example | • 2-74: | Designer: For example, you can see the corresponding circuit diagram after connecting the physical diagram, or vice versa |
| | • 2-81: | Designer: For example, see the circuit diagram to connect the physical diagram, or vice versa |
| provide feedback | • 2-70: | User: This is good, I hope it can be very convenient to draw circuit diagrams, and I also hope there is a function to hide the background grid with one click, which is convenient for us teachers to take screenshots for test papers. Currently, there is no software that makes it easy to draw circuit diagrams for exam questions |
| | • 2-85: | User: Indeed |
| | • 2-93: | User: Yes, the common one is still the old style |
| | • 2-126: | User: Updated |
| provide information | • 2-34: | Designer: Touch screens have their own on-screen keyboards. |
| provide positive feedback | • 2-78: | User: The 3D effect of your software is very good |
| | • 2-89: | User: This idea is really good |
| provide practical advice | • 2-36: | Designer: Last time I tried it in middle school, there's a trick where you click the link button in the top right, and when the browser pops up, it's no longer full screen... = = |
| provide progress report | • 2-61: | Designer: [Image] Updates in preparation |
| provide release schedule | • 2-105: | Designer: There will be an update. Submitting to the app store / releasing the Android version next Monday, and releasing the Windows version over the weekend. |
| provide solution for download query | • 2-53: | Designer: Group sharing. |
| provide specific feedback | • 2-96: | User: Left side |
| provide technical details | • 2-16: | Designer: The student power supply supports both DC and AC and is an ideal component. |
| provide technical timeframe | • 2-28: | Designer: Theoretically, it should be compatible (but the machine itself shouldn't be too old, probably from 2008 onwards). |
| provide timeline | • 2-62: | Designer: This is quite complex, so it will take more time... Hopefully, it can be released this week |
| provide usage instruction | • 2-124: | Designer: After connecting the student power supply, you need to turn on the switch |
| provide visual aid | ● 2-91: | Designer: Consulting the teachers in the group: which type of intersection is used in the circuit diagrams in the current textbooks? [Image] |
| | ● 2-92: | Designer: [Image] This one, right? |
| provide workaround | ● 2-101: | Designer: This version of the circuit diagram is for testing only... You can drag the Editor because the auto-layout algorithm is not very stable and may produce some odd results |
| | ● 2-110: | Designer: Although you can manually adjust the layout |
| reassure user | ● 2-23: | Designer: 7 is okay. |
| | ● 2-39: | Designer: There will be. |
| reassure users | ● 2-109: | Designer: Don't worry, don't worry, it will come, just optimizing the circuit diagram one last time. |
| reciprocate greeting | ● 2-59: | Designer: Hello :) |
| recognize usability limitation | ● 2-35: | Designer: But you can't bring it up without touching the input, that's probably the case. |
| reference past experience | ● 2-36: | Designer: Last time I tried it in middle school, there's a trick where you click the link button in the top right, and when the browser pops up, it's no longer full screen... = = |
| reference past incident | ● 2-41: | User: Generally not. Ever since an adult image popped up during a major city-level open class, the school has disabled the network on classroom computers [Emoji]. |
| reiterate bidirectional conversion feature | ● 2-81: | Designer: For example, see the circuit diagram to connect the physical diagram, or vice versa |
| reiterate lack of internet | ● 2-44: | User: Most schools will disable the network on classroom computers to prevent students from going online or to avoid various software auto-downloads from slowing down the computers. |
| reiterate need for user manual | ● 2-68: | User: Need the user manual for this software |
| request additional feature | ● 2-73: | User: Can you also include mechanics experiments? |
| request private message | ● 2-69: | Designer: Oh, please message me privately and tell me what you roughly need |
| request specific features | ● 2-8: | User: First, let's pay homage to the experts, then I'll make a small request. Could you create a dynamic demonstration of mechanical waves and mechanical vibrations, such as the propagation of transverse and longitudinal waves, wave superposition, diffraction, and interference? Also, for optical experiments, it would be great to have optical benches, single slits, double slits, and polarizers to demonstrate optical experiments. |

| request specific feedback | • 2-94: | Designer: Uh... left side or right side |
|---|---|---|
| request suggestions | • 2-17: | Designer: Okay. Please give more suggestions! |
| request technical help | • 2-1: | Designer: @Morning Tea Moonlight How can I upload a high-definition, uncensored version of this crappy avatar? |
| seek attention | • 2-52: | User: Excuse me. |
| seek clarification | • 2-92:<br>• 2-94: | Designer: [Image] This one, right?<br>Designer: Uh... left side or right side |
| seek consensus | • 2-100: | Designer: So, use both? |
| seek guidance | • 2-64: | User: Is there a user manual? |
| seek guidance on pc version | • 2-18: | User: How do you use the PC version? |
| seek help | • 2-51: | User: How to download? |
| seek specific platform information | • 2-2: | User: PC version? |
| seek to improve visual representation | • 2-1: | Designer: @Morning Tea Moonlight How can I upload a high-definition, uncensored version of this crappy avatar? |
| set development goal | • 2-77: | Designer: Hope to figure out electromagnetism before the end of the year |
| set expectations | • 2-28: | Designer: Theoretically, it should be compatible (but the machine itself shouldn't be too old, probably from 2008 onwards). |
| share ongoing research | • 2-71: | Designer: Actually, we are researching the bidirectional conversion between physical diagrams and circuit diagrams |
| share positive classroom experience | • 2-46: | User: The simulation effect of this software is really good. When I used it in class yesterday, the students were amazed. |
| share progress | • 2-29: | Designer: Hello everyone~ Version 1.01 has been released in the group files, the updates are as follows: |
| share teaching experience | • 2-48: | User: I taught high school for seven years, and now I've been teaching middle school for eight years. |
| share visual update | • 2-61: | Designer: [Image] Updates in preparation |
| share workaround | • 2-36: | Designer: Last time I tried it in middle school, there's a trick where you click the link button in the top right, and when the browser pops up, it's no longer full screen... = = |
| show anticipation | • 2-104:<br>• 2-125: | User: Will there be an update this week? [Emoji]<br>User: Has the Android version not been updated yet? |

| show appreciation | ● 2-11: | User: Okay, okay~ Thank you for your hard work. |
|---|---|---|
| show consideration for users | ● 2-50: | Designer: Sorry for the late-night disturbance :) |
| show interest in development progress | ● 2-60: | User: Are there any other updates recently? |
| show interest in user context | ● 2-47: | Designer: Haha. Are you a middle school or high school teacher? |
| show interest in using the software | ● 2-51: | User: How to download? |
| show politeness | ● 2-52: | User: Excuse me. |
| show respect to experts | ● 2-8: | User: First, let's pay homage to the experts, then I'll make a small request. Could you create a dynamic demonstration of mechanical waves and mechanical vibrations, such as the propagation of transverse and longitudinal waves, wave superposition, diffraction, and interference? Also, for optical experiments, it would be great to have optical benches, single slits, double slits, and polarizers to demonstrate optical experiments. |
| solicit feedback | ● 2-91: | Designer: Consulting the teachers in the group: which type of intersection is used in the circuit diagrams in the current textbooks? [Image] |
| solicit user feedback | ● 2-55: | Designer: We will also update the multimeter and powered solenoid this week. If there are any other features, characteristics, or electronic components you hope to include in the production plan, please leave a message in the group, and we will consider it as much as possible. The previously mentioned feature of saving experiments to the cloud/local is also in the plan. Thank you, everyone! |
| suggest alternative system | ● 2-22: | User: Win7 should be fine. |
| suggest broader usability feature | ● 2-33: | Designer: Speaking of which, doesn't that mean every place where numbers are input should have a soft keyboard? |
| suggest expansion to mechanics | ● 2-73: | User: Can you also include mechanics experiments? |
| suggest group sharing | ● 2-53: | Designer: Group sharing. |
| suggest manual adjustment | ● 2-110: | Designer: Although you can manually adjust the layout |
| suggest new feature | ● 2-31: | User: It is recommended to add an exit button function to the PC version. Many regions now use all-in-one touch screen machines without physical keyboards. |

| suggest new functionality | ● 2-38: | User: If there could be an export function, or the ability to save or import experiments, it would be convenient. We could set up the parameters in the office and directly import them in class. |
|---|---|---|
| suggest potential future project | ● 2-87: | Designer: We'll see if there's a suitable opportunity to do one later |
| suggest potential solution | ● 2-27: | Designer: XP compatibility might need to be looked into later... probably need to install a virtual machine. |
| suggest prioritization | ● 2-82: | User: Don't aim for completeness, it should be categorized and refined one by one |
| suggest simplification | ● 2-98: | Designer: Make it simpler... |
| theorize compatibility | ● 2-28: | Designer: Theoretically, it should be compatible (but the machine itself shouldn't be too old, probably from 2008 onwards). |
| update community | ● 2-15: | Designer: From the next update, dry batteries will no longer be ideal components (you can remove the internal resistance to simulate). |
| update on android version timeline | ● 2-123: | Designer: The Android version is expected to update tonight |
| update on development progress | ● 2-55:<br><br><br><br><br><br>● 2-63: | Designer: We will also update the multimeter and powered solenoid this week. If there are any other features, characteristics, or electronic components you hope to include in the production plan, please leave a message in the group, and we will consider it as much as possible. The previously mentioned feature of saving experiments to the cloud/local is also in the plan. Thank you, everyone!<br>Designer: There will be: multimeter; powered solenoid; semiconductor capacitor; support for conversion to ideal ammeter (more convenient for problem-solving and middle school teaching) |
| update on feature development | ● 2-71: | Designer: Actually, we are researching the bidirectional conversion between physical diagrams and circuit diagrams |
| update on progress | ● 2-95:<br>● 2-117: | Designer: We are working on this part<br>Designer: The new version has already been sent in the group |
| update on release process | ● 2-114: | Designer: The Apple Store is still reviewing, let's upload the PC version first |
| use emoji to convey sentiment | ● 2-80: | User: It's already starting to take shape [Emoji] |

| | |
|---|---|
| use emoji to greet | ● 2-12:   User: [Emoji] Hello everyone, may I ask where I can download the PC / interactive whiteboard version of Physics Lab? |
| use humor | ● 2-41:   User: Generally not. Ever since an adult image popped up during a major city-level open class, the school has disabled the network on classroom computers [Emoji]. |
| use non verbal communication | ● 2-57:   User: [Emoji][Emoji] |
| user engagement | ● 2-104:  User: Will there be an update this week? [Emoji]<br>● 2-125:  User: Has the Android version not been updated yet? |
| validate experience | ● 2-37: User: @Designer Yes, yes. |